# Physics-based Prediction, uncertainty quantification and decision-making for IN718 crystallographic texture intensity across LPBF defocus regimes

**Authors:**

Yisheng Lu[1], John Riris[2], Jie Song[2], Yao Fu[2,3], Jie Chen[1,3,4]

**Affiliations:**

[1]Department of Mechanical Engineering, Virginia Tech, Blacksburg, VA 24061, USA

[2]Department of Aerospace & Ocean Engineering, Virginia Tech, Blacksburg, VA 24061, USA

[3]VT Made, Virginia Tech, Blacksburg, VA 24061, USA

[4]Macromolecules Innovation Institute, Virginia Tech, Blacksburg, VA 24061, USA

## Abstract

Reliable prediction of crystallographic texture in laser powder bed fusion is critical for linking process conditions with anisotropic response and for qualification. However, black-box models may fail under shift and cannot distinguish weak data support from loss of physical validity. This study develops a two-stage physics-based model for <001> || BD (build direction) texture in Inconel 718. Stage 1 maps process variables to melting mode and melt pool geometry. Stage 2 predicts texture by combining an empirical physics model with a random-forest residual model. A k-nearest-neighbor weight attenuates residual corrections for poorly supported queries, while a study-specific areal beam-power-density criterion withholds predictions outside the adopted conduction envelope. Conformal intervals are evaluated on the retained physics-valid set, and SHAP and Sobol analyses assess residual sensitivity. Under a controlled leave-one-defocus-out evaluation, the physics anchor achieved $R^2$ = 0.778, against -0.001 for the black-box model and 0.750 for the gated hybrid. Under leave-one-group-out cross-validation, the gated hybrid reached $R^2 = 0.592$ against 0.538 for the black-box model. Retained set coverage was 92.9% at a mean full width of 3.65 multiples of a uniform distribution (MUD) under grouped cross-validation and 100% at a width of 3.21 MUD under transfer to a withheld +80 mm defocus regime. An illustrative mapping produced a retained BD elastic-modulus span of 127-187 GPa. On nine conditions from a separately built sample set, the framework withheld three, attenuated three, and matched the measured ordering for the rest. Separating data applicability, physics validity, and predictive uncertainty into distinct decisions lets the framework transfer where an unconstrained model does not, and withhold predictions where no model class performs adequately.

## 1. Introduction

Laser powder bed fusion (LPBF) enables the fabrication of geometrically complex metallic components through localized melting and layer wise consolidation. However, the steep and directionally varying thermal gradients generated during LPBF also produce heterogeneous solidification conditions and crystallographic textures [1,2]. In face-centered-cubic nickel-based alloys, elastic stiffness is strongly orientation dependent, with ⟨001⟩ more compliant than ⟨111⟩ [3–5]. During LPBF, competitive columnar growth along a predominantly build-aligned thermal gradient can preferentially align ⟨001⟩ with the build direction (BD). Quantifying this enrichment therefore provides a compact texture target for linking process conditions to direction-dependent material response [9-11]. Predicting this texture from controllable process variables is therefore important for connecting process design with the expected material response [6,7].

Crystallographic texture is governed by a chained process-to-melt-pool-to-solidification relationship [2,8]. Laser power (P), scan speed (V), hatch spacing (H), and focus offset (F) determine the local energy deposition and thermal history. These conditions influence melt-pool geometry, overlap between adjacent tracks, repeated remelting, and the thermal-gradient orientation associated with competitive grain selection [8–11]. Shao et al. [11] demonstrated that hatch spacing and laser remelting reorganize the ⟨001⟩ texture of IN718 in LPBF, showing that texture formation is a history-dependent, multi-cycle process rather than a single-track outcome. Transitions between conduction- and keyhole-dominated melting further alter melt-pool geometry and the associated solidification conditions [9,12–14]. From an engineering standpoint, texture intensity should not be evaluated independently of melt-pool stability. Keyhole-mode melting produces vapor depressions, and instability or collapse of this cavity can promote pore formation [12,14]. The present study therefore treats an adopted conduction-dominated processing envelope as its intended deployment domain, rather than treating high texture intensity obtained under any melting state as equally desirable. Scipioni Bertoli et al. [15] showed that volumetric energy density (VED) alone cannot represent these coupled effects, because different process-parameter combinations producing the same nominal energy input generate different beam sizes and melt-pool states.

Data-driven models offer an efficient alternative to repeated experiments and high-fidelity thermal-fluid simulations. Machine-learning models have been applied to melt-pool geometry prediction and in situ or post-build defect classification in metal additive manufacturing [16–20] and comparatively fewer studies have linked LPBF process conditions to crystallographic texture or developed machine-learning surrogates for texture evolution [21,22]. Tree-based and Gaussian-process formulations allow nonlinear process interactions to be learned from limited observations. However, predictive performance within randomly partitioned data does not establish transferability to unobserved process regimes: a black-box model can achieve favorable interpolation accuracy while producing unsupported corrections once a query lies beyond its training distribution. Physics-informed and hybrid formulations address this limitation by constraining the learned relationship or anchoring it to a physically interpretable trend [23], and a recent review [24] surveyed these strategies for process-structure-property modeling in additive manufacturing. Their behavior under controlled regime extrapolation, however, still requires explicit evaluation.

Furthermore, reliability assessment should also account for the structure of the validation data. Roberts et al. [25] showed that when data have a grouped structure, random sample-level splitting places the closely related conditions in both training and testing subsets, and yields optimistic performance estimates, so group-aware validation is required when the intended use involves prediction for unseen groups. Complementary to grouping, Sutton et al. [26] introduced applicability-domain analysis to assess whether a query is supported by the training distribution of a materials-science model, a notion later generalized to distance-based support measures [27]. For quantifying predictive uncertainty, Lei et al. [28] established distribution-free regression intervals under exchangeability, providing the statistical basis for conformal calibration. Subsequent work has shown that conformal validity under distribution shift remains conditional on the calibration design and exchangeability assumptions [29,30], with conditional and hierarchical guarantees requiring additional structure [31–33]. Therefore, a reliable workflow should distinguish residual-model applicability, physics-domain validity, and uncertainty calibration rather than collapsing them into a single confidence measure.

A remaining need is an integrated model that connects process conditions to texture while preserving these distinctions. The intermediate melt-pool state provides a physically meaningful bridge, but surrogate errors can propagate into texture prediction. In this study, a process-regime shift refers to a change in the focus-offset level and the associated surface beam radius rather than a change in melting mode. Under such a shift, the framework must determine whether the learned residual remains supported, whether the conduction-inspired anchor remains valid, and whether an uncertainty interval can be issued. These requirements favor separate mechanisms for applicability attenuation, physics-validity abstention, and retained-set uncertainty.

In this study, we develop a two-stage reliability-oriented framework for predicting ⟨001⟩ ∥ BD texture enrichment within an adopted conduction-dominated envelope. Stage-1 maps the LPBF process variables to printability, melting mode, and melt-pool geometry, with its predicted depth and width serving as fixed model-derived mediators. Stage-2 combines a conduction-inspired greybox anchor with a random forest (RF) residual model. A standardized k-nearest-neighbor (kNN) distance attenuates the residual as data support weakens, while a separate, study-specific areal beam-power-density criterion withholds predictions outside the adopted physics-validity envelope. Conformal intervals are calibrated through grouped cross-validation (group-CV) and reported only for retained predictions.

The main contributions of this study are summarized as follows.

1. The proposed framework separates data applicability, physics validity, and predictive uncertainty, which are often combined into a single confidence measure. The learned correction is attenuated as local data support weakens, predictions are withheld when the anchor lies outside its adopted validity envelope, and intervals are reported only for predictions that are issued.

2. The evaluation is designed to assess transfer across experimental groups and process regimes. Complete experimental groups and defocus levels are excluded from model fitting, and literature-derived model forms are reimplemented on the same dataset under the same evaluation protocol to provide a consistent comparison.

3. The resulting reliability behavior is demonstrated through withheld-regime evaluation and separately built specimens not used in model development. The physics-anchored framework retains predictive capability in a withheld defocus condition where the unconstrained data-driven model does not, and issues nothing in a regime where no evaluated model class performs adequately. On the separately built set the framework applies the same distinction without adjustment, withholding, attenuating, and issuing predictions according to its own criteria.

The remainder of this paper is organized as follows. Section 2 describes the experimental setups and datasets, and section 3 introduces two-stage framework, validation protocol, and statistical analyses. Section 4 presents and discusses predictive performance, transfer and abstention behavior, uncertainty results, and model interpretation. Section 5 illustrates the propagation of retained texture predictions to build-direction modulus, applies the frozen framework to a separately built set, and examines the assumed orientation dependence at room and elevated temperature, followed by the conclusions in Section 6.

## 2. Experimental setups and datasets

All specimens were fabricated from Inconel 718 powder with a near-normal particle size distribution centered at approximately 30 µm, using a Nikon SLM 280 LPBF system equipped with a Ytterbium continuous-wave fiber laser (nominal wavelength of 1070 nm, nominal spot size of 80 ± 10 µm, Rayleigh length 4 ± 1 mm, beam quality $M^2 = 1.0–1.5$, Gaussian beam profile [34,35]) under an argon atmosphere. A meandering scan strategy with 90° inter-layer rotation was applied, and the layer thickness was held constant at 30 µm throughout. The process space was designed to span conduction-dominated, transitional, and keyhole-dominated melting conditions. Melt-pool geometry was measured on etched cross-sections prepared perpendicular to the scanning direction, with five melt pools evaluated per parameter combination. Tracks showing severe lack of fusion or discontinuous melt-pool formation were labelled as failed. Each successful track was assigned to conduction or keyhole mode from the observed cross-sectional morphology. Only width and depth were recorded for conduction-mode pools, while maximum width, overall depth, transition width and transition depth were for keyhole-mode pools.

Two linked datasets supported the process-to-melt-pool-to-texture workflow. Stage-1 comprised 372 conditions spanning laser powers of 150–400 W, scan speeds of 75–2,000 mm $s^{-1}$, hatch spacings of 50–400 µm, and focus offsets of −160 to +80 mm, with the sign convention defined in Figure 1a. Of these, 345 conditions were successful and 27 failed. Printability classification used all 372 conditions, whereas melting-mode classification and geometry regression used the 345 successful conditions, comprising 257 conduction-mode (CM) and 88 keyhole-mode (KM) cases. The focus-offset convention and its effect on the surface beam radius are illustrated in Figure 1a. Stage-1 data were used only for the process-to-melt-pool bridge described in Section 3.2.

The Stage-2 dataset comprised 278 electron backscatter diffraction (EBSD) observations across nine predefined experimental groups, designated G1 through G9 (Table 1). These observations represented 269 unique process vectors (P, V, H, F); nine vectors occurred in two groups and were retained as distinct group-level texture measurements.

**Table 1.** Composition of the Stage-2 dataset across the nine predefined experimental groups

| Experimental group | *n* | Focus-offset composition (mm: N) | P (W) | V (mm $s^{-1}$) | H (µm) |
|---|---|---|---|---|---|
| G1 | 38 | 20 (38) | 200–400 | 100–600 | 100–400 |
| G2 | 36 | 0 (12); 80 (24) | 250–400 | 100–600 | 100–500 |
| G3 | 41 | 0 (11); 40 (30) | 150–400 | 100–800 | 75–400 |
| G4 | 30 | 40 (18); 80 (12) | 300–400 | 200 | 125–350 |
| G5 | 42 | 0 (12); 20 (24); 40 (6) | 150–400 | 200 | 50–300 |
| G6 | 18 | 0 (4); 20 (3); 40 (7); 80 (4) | 400 | 400–1000 | 50–175 |
| G7 | 21 | 0 (15); 20 (6) | 400 | 1000–1600 | 25–125 |
| G8 | 11 | 20 (6); 40 (3); 80 (2) | 400 | 400–1200 | 25–125 |
| G9 | 41 | 10 (18); 20 (11); 30 (12) | 400 | 500–700 | 50–225 |
| **Overall** | **278** | **0: 54; 10: 18; 20: 88; 30: 12; 40: 64; 80: 42** | **150–400** | **100–1600** | **25–500** |

*Groups are designated G1–G9; the correspondence with the acquisition identifiers used in the distributed project archive is given in the Supplementary Information. Focus-offset entries give the focus value (mm) followed by the observation count in parentheses. The nine groups define the leave-one-group-out (LOGO) units. The +80 mm (n = 42) and focus-zero (n = 54) holdouts span multiple groups and are defined at the dataset level.*

The two datasets shared overlapping process ranges but supported different response variables. Stage 1 used melt-pool labels to learn printability, melting mode, and geometry, whereas Stage 2 used the corresponding process variables together with the Stage-1 predicted geometry to model crystallographic texture. No measured Stage-1 geometry label was used directly as a Stage-2 input, and the Stage-1 models were fitted to the complete melt-pool dataset before generating the model-derived depth and width for the 278 EBSD observations. Stage-1 extended farther in scan speed and focus offset, while Stage-2 covered a wider hatch-spacing range. Fourteen Stage-2 observations lay outside the Stage-1 hatch-spacing range of 50–400 µm. They were retained and flagged as Stage-1 hatch-range extrapolations, so their geometry estimates represent boundary-supported tree-model predictions rather than interpolation.

Tensile specimens were built as cylindrical bars and machined on a CNC lathe to a nominal gauge length of 20 mm and a gauge diameter of 4.75-5.0 mm, with threaded sections at both ends. The tensile axis was parallel to the build direction.

Specimens for texture characterization were taken from a build height of approximately 13 mm and sectioned so that the observation-plane normal corresponded to the build direction. The specimens were mounted in conductive resin and mechanically polished to a 0.05 μm colloidal silica finish without subsequent etching. Crystallographic texture was characterized by electron backscatter diffraction (EBSD) using a Thermo Scientific Helios 5 UC DualBeam FIB-SEM equipped with an EDAX EBSD system at the Nanoscale Characterization and Fabrication Laboratory, Virginia Tech. Scans were acquired at an accelerating voltage of 30 kV using a hexagonal grid with a step size of 1.2 to 2 μm over areas of approximately 0.8 × 0.6 to 1.0 × 0.8 mm, each containing hundreds of grains. Indexing rates exceeded 99%, and the indexed points were treated as the valid orientation points from which the texture target was constructed. Orientation data were processed using TSL OIM Analysis and the MTEX toolbox.

Each Stage-2 observation was associated with a process vector, an experimental-group label, and an EBSD-derived crystallographic texture target. The primary target was the enrichment of ⟨001⟩ orientations parallel to the build direction (BD). Let $f_{001}$ denote the percentage of valid indexed orientation points for which any symmetry-equivalent ⟨001⟩ direction lay within $15^{\circ}$ of BD. The texture target was defined as the ratio of $f_{001}$ to the adopted random-texture reference fraction of 10.21%:

$$e_{001} = \frac{f_{001}}{10.21}. \tag{1}$$

The normalized target $e_{001}$ was expressed in multiples of a uniform distribution (MUD), where $e_{001}=1$ represents the random texture and values above unity indicate ⟨001⟩ ∥ BD enrichment. The dataset spanned 0.533–7.965 MUD. This point-fraction-based target was determined solely from the measured EBSD orientations and was fixed before model fitting. Pole-figure maxima, composite texture scores, and model-derived quantities were excluded. Figure 1b, c illustrate lower and higher enrichment.

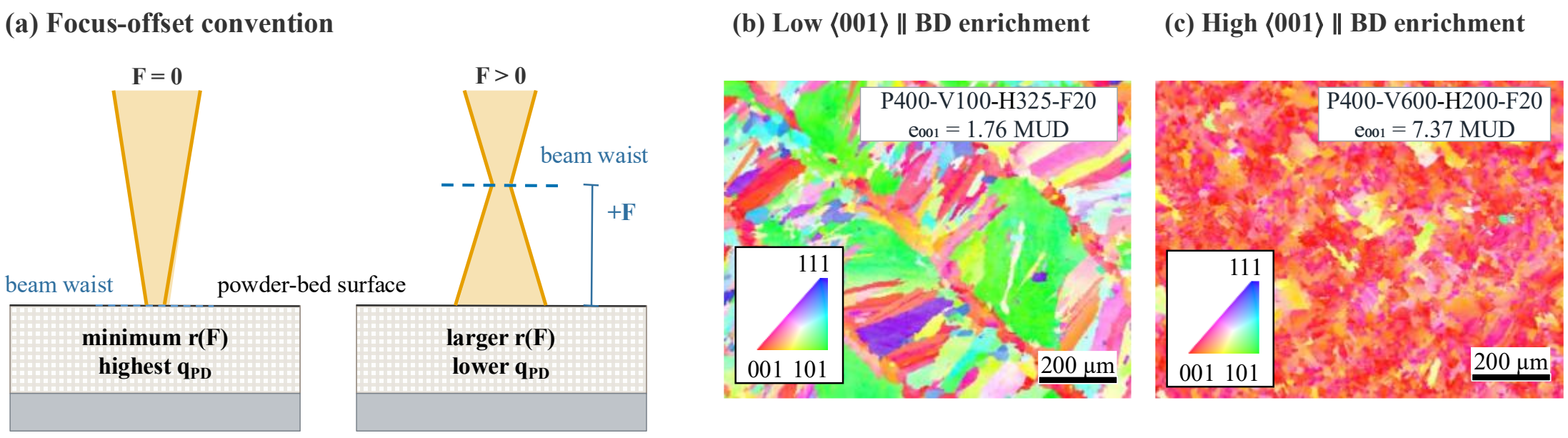

**Figure 1.** Focus-offset convention and the EBSD-derived texture target. (a) At (F=0), the beam waist lies at the powder-bed surface; (F>0) moves the focal plane toward the laser, increasing r(F) and reducing $q_{PD}$. Beam geometry is not to scale. (b, c) BD-referenced inverse pole figure maps for two observations from G1 with lower and higher ⟨001⟩ ∥ BD enrichment. The target $e_{001}$ is the fraction of valid indexed points within 15° of ⟨001⟩ ∥ BD, normalized by the 10.21% random-texture fraction; it is not derived from pole-figure intensity.

The nine groups were retained as the units for leave-one-group-out (LOGO) validation and conformal calibration. Because nine process vectors occurred in two groups, a held-out group could share a process vector with the training groups. LOGO therefore evaluates transfer to an unseen experimental group rather than strict leave-one-process-condition-out extrapolation.

# 3. Physics-based machine learning and statistical analysis

## 3.1 Overall physics-based modeling framework

The proposed framework predicts the ⟨001⟩ ∥ BD texture intensity from LPBF process parameters through a two-stage predictor and a separate reliability layer (Figure 2). Stage-1 maps P, V, H, and F, together with their derived descriptors, to printability, melting mode, and melt-pool depth and width. The predicted geometry is passed to Stage-2 as model-derived mediators. Stage-2 combines a conduction-inspired greybox anchor with an RF residual correction. A standardized kNN distance attenuates the residual as empirical support weakens, while a separate areal beam-power-density criterion withholds predictions and intervals outside the adopted anchor-validity envelope. For retained predictions, group-CV conformal calibration provides empirical 90% intervals. Residual-model applicability, physics-domain validity, and uncertainty calibration therefore remain separate decisions. Sections 3.2-3.4 detail the Stage-1 bridge, Stage-2 predictor, and reliability layer, respectively.

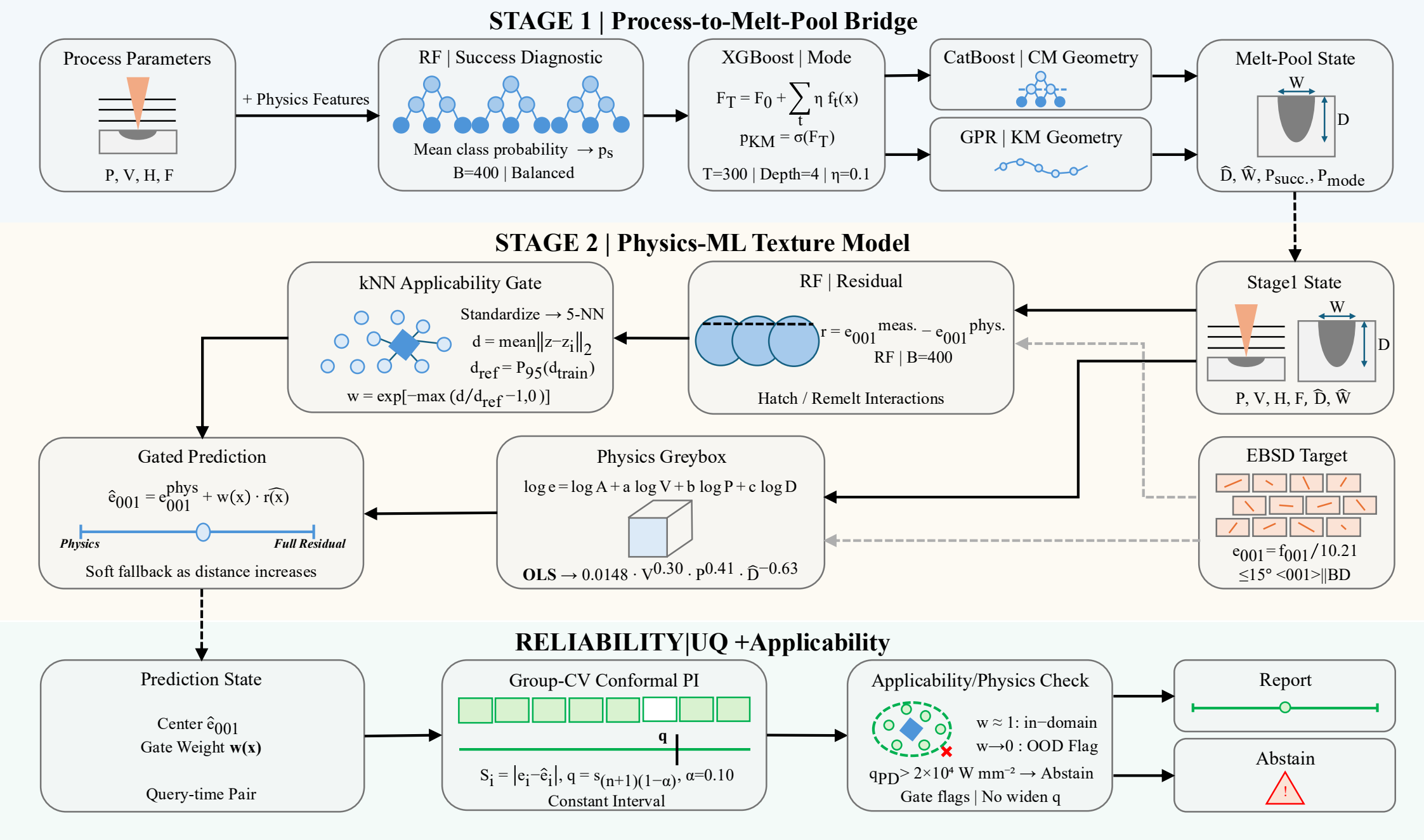

**Figure 2.** Overall physics-based framework for prediction and uncertainty quantification of IN718 crystallographic texture under process-regime shift in LPBF. Stage-1 predicts printability, melting mode, and melt-pool geometry. Stage-2 combines a conduction-inspired greybox anchor with a random-forest residual correction. Data-support weighting attenuates the residual, physics-validity screening controls abstention, and grouped conformal calibration provides empirical 90% intervals for retained predictions. Solid arrows show prediction flow; dashed arrows connect measured targets used for training or calibration.

### 3.2 Stage-1 process-to-melt-pool bridge

Eight descriptors were derived from the four process variables using fixed material and optical constants: six energy- and transport-related proxies and two focus-offset descriptors, |F| and sign (F). The effective Gaussian beam radius was calculated as

$$r(F)=w_0\sqrt{1+\left(\frac{F}{z_R}\right)^2} \tag{2}$$

where $w_0$=0.040 mm and $z_R$=4.0 mm denote the beam-waist radius and Rayleigh length, respectively. The remaining energy and transport descriptors were calculated as

$$E_L = \frac{P}{V} \tag{3}$$

$$E_V = \frac{P}{V\,H\,t} \tag{4}$$

$$q_{PD} = \frac{P}{\pi r(F)^2} \tag{5}$$

$$C_{proxy} = \frac{V}{E_V} \tag{6}$$

$$Pe = \frac{(V/1000)\,[2r(F)/1000]}{\alpha_{th}} \tag{7}$$

$$\Pi_E = \frac{P}{V\,H} \tag{8}$$

The six derived descriptors comprise line energy $E_L$ (J mm$^{-1}$), volumetric energy $E_V$ (J mm$^{-3}$), the areal beam-power-density proxy $q_{PD}$ (W mm$^{-2}$), the areal-energy proxy $\Pi_E$ (J mm$^{-2}$), the dimensionless Péclet number Pe, and $C_{proxy}$, an empirical cooling proxy retained in its implemented form with derived units of mm$^4$ J$^{-1}$ s$^{-1}$. H was converted from μm to mm. The feature calculation used $t=0.030$ mm and $\alpha_{th}=5.3\times10^{-6}$ m$^2$ s$^{-1}$ for the beam-waist radius and Rayleigh length of the installed fiber laser, taken from the manufacturer specification, and a thermal diffusivity of $5.3 \times 10^{-6}$ m$^2$ s$^{-1}$ for IN718 near the liquidus [36]. The descriptors $|F|$ and $\mathrm{sign}(F)$ preserved both the magnitude and direction of defocus. Together with the original process variables P, V, H, and F, these descriptors formed the 12-dimensional Stage-1 input vector:

$$\mathbf{x}_{MP} = [P, V, H, F, |F|, \mathrm{sign}(F), E_L, E_V, q_{PD}, C_{proxy}, Pe, \Pi_E] \tag{9}$$

Stage-1 used three sequential branches. A balanced RF first classified whether a condition produced a successful melt pool [37]. Successful conditions were then routed by an XGBoost keyhole-probability classifier, with a threshold of 0.5 assigning each query to the CM or KM geometry branch [38]. CatBoost predicted CM depth and width [39], while Gaussian-process regression (GPR) predicted KM depth and maximum width after feature standardization and response normalization [40]. Model configurations and validation results are summarized in Table 2. All stochastic Stage-1 models used a fixed random seed of 42.

**Table 2.** Stage-1 model configurations and validation results for printability, melting-mode routing, and mode-specific geometry regression.

| Task | Deployed model | Validation | Result |
|---|---|---|---|
| Printability | Balanced RF (400 trees) | 5-fold stratified CV | ROC-AUC = 0.802; accuracy = 0.952 |
| CM/KM classification | XGBoost (300 trees; depth 4; learning rate 0.1) | 5-fold stratified CV | ROC-AUC = 0.990; accuracy = 0.954 |
| CM geometry | CatBoost (300 iterations; depth 4; learning rate 0.1) | 5-fold GroupKFold by (P, V, H, F) | Depth $R^2$ = 0.849; width $R^2$ = 0.868 |
| KM geometry | Standard Scaler + GPR constant × radial basis function + white-noise kernel | 5-fold GroupKFold by (P, V, H, F) | Depth $R^2$ = 0.781; width $R^2$ = 0.803 |

Geometry models were evaluated by five-fold GroupKFold using the complete process vector (P, V, H, F) as the grouping key. Performance was summarized by the coefficient of determination ($R^2$) and root-mean-square error (RMSE). Because each of the 345 successful Stage-1 conditions was unique, GroupKFold was equivalent to a condition-level partition and could not estimate repeated-condition or repeated-build variability.

After validation, the Stage-1 models were refitted on the complete melt-pool dataset and applied to the 278 Stage-2 observations. CM-routed queries received CatBoost depth and width predictions, whereas the KM-routed query received GPR depth and maximum-width predictions. The resulting $\hat{D}$ and $\hat{W}$ values served as model-derived Stage-2 mediators

All 278 Stage-2 observations had measured EBSD targets and were therefore retained for texture modeling. The Stage-1 printability probability for these observations was recorded as a diagnostic output and was not used to exclude observations.

Meanwhile, the Stage-1 classifier served to select the CM or KM geometry surrogate used to generate $\hat{D}$ and $\hat{W}$. It routed 277 observations to the CM branch and one to the KM branch. Consequently, the model-derived mediators entering Stage 2 were generated almost entirely by the CM geometry surrogate, providing context for the conduction-inspired form of the Stage-2 anchor. Branch assignment did not establish the physics validity of that anchor or determine whether a Stage-2 prediction was issued. Prediction issuance was assessed separately using the $q_{PD}$ criterion defined in Section 3.4.

Figure 3 and Table 2 summarize Stage-1 validation. Depth parity is shown because $\hat{D}$ enters the Stage-2 greybox directly, whereas $\hat{W}$ enters only the residual model. Grouped-CV RMSEs for CM and KM depth were 0.021 and 0.091 mm respectively, and the corresponding width RMSEs were 0.041 and 0.081 mm.

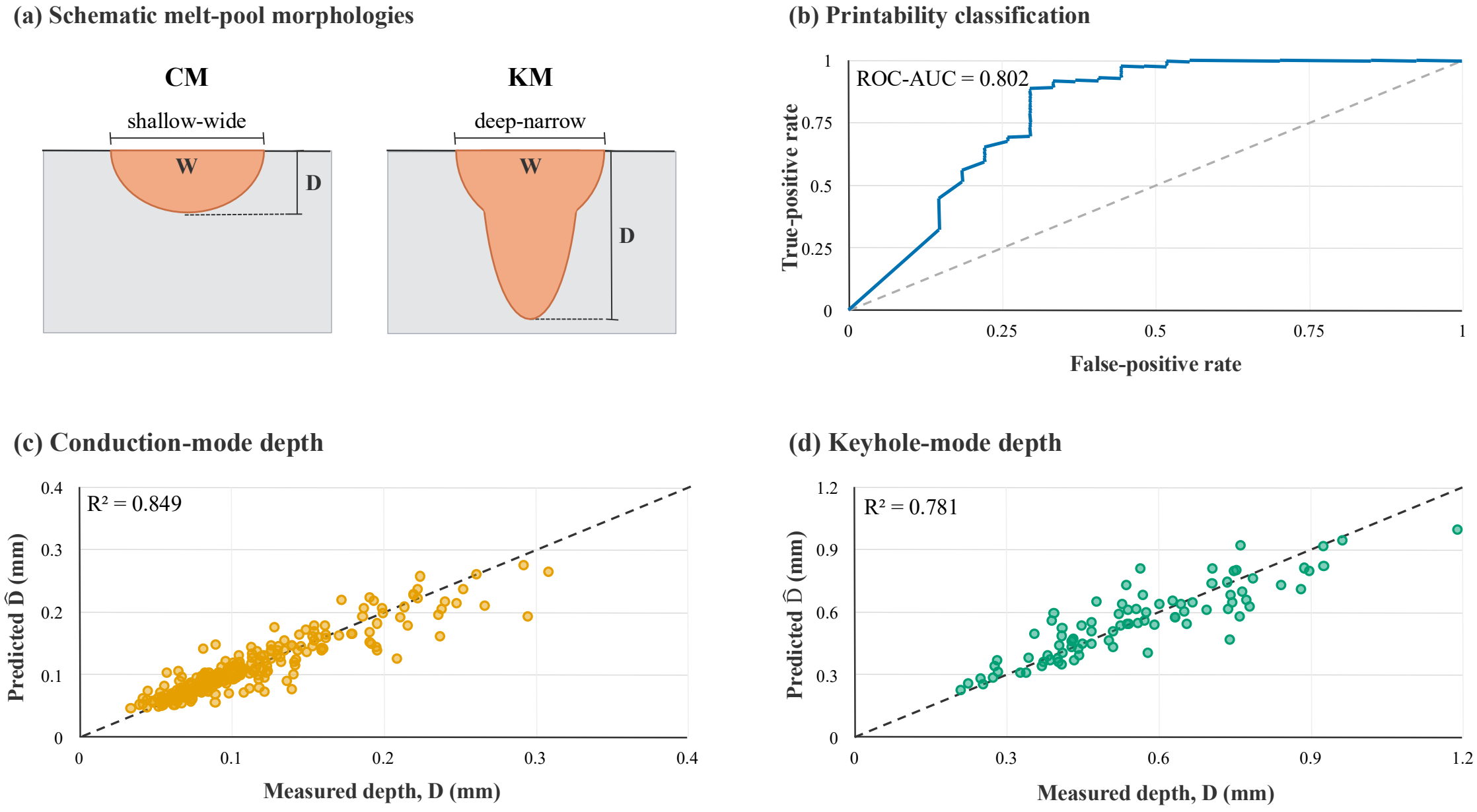

**Figure 3.** Physical labels and validation of the Stage-1 bridge. (a) Schematic conduction-mode (CM) and keyhole-mode (KM) melt-pool cross sections. (b) Five-fold stratified-CV receiver operating characteristic for the balanced random-forest printability classifier. (c, d) Grouped-CV depth parity for the CM CatBoost and KM Gaussian Process regression models respectively.

### 3.3 Stage-2 physics-anchored residual model and applicability weighting

Stage-2 predicted the normalized texture target $e_{001}$ from the process variables and Stage-1 melt-pool mediators. It combined a conduction-inspired greybox anchor with a learned statistical discrepancy [41]. For observation i, the deployed six-component input vector was

$$\mathbf{x}_i=[P_i, V_i, H_i, F_i, \hat{D}_i, \widehat{W}_i] \tag{10}$$

where $\hat{D}_i$ and $\widehat{W}_i$ are the mode-specific depth and width predictions generated by the Stage-1 surrogate. For KM-routed queries, $\widehat{W}_i$ denoted the predicted maximum width. The melting-mode label selected the Stage-1 geometry branch but was not included as a separate Stage-2 input

A positive multiplicative form in V, P, and $\hat{D}$ was adopted as a low-capacity empirical anchor.

Laser power and scan speed represent the primary energy-input and interaction-time variables, whereas $\hat{D}$ supplies a model-derived melt-pool-state mediator that carries the thermal-gradient information not recoverable from the process variables alone. The fitted exponents were treated as empirical associations in the stated units rather than as universal physical constants. Recent operando evidence indicates that local abnormal columnar-to-equiaxed transition in metal additive manufacturing may involve ordering-mediated nucleation pathways not resolved by classical criteria based solely on thermal gradient and growth rate [42]. Although the present anchor is not formulated in terms of these quantities, this finding provides broader context for interpreting it as a compact empirical association rather than a complete mechanistic model of texture formation. The greybox anchor related $e_{001}$ multiplicatively to V, P, and $\hat{D}$ and was fitted by ordinary least squares after logarithmic transformation:

$$\log e_{001,i}=\log A+a\log V_i+b\log P_i+c\log \hat{D}_i+\varepsilon_i \quad (11)$$

The full-data fit used for final deployment was

$$e_{001,i}^{phys}= 0.0148\, V_i^{0.30}P_i^{0.41}\hat{D}_i^{-0.63} \quad (12)$$

Here, P, V, and $\hat{D}$ were expressed in W, mm $s^{-1}$, and mm, respectively, and $\hat{D}$ was a Stage-1 prediction rather than a measured quantity. Equation 12 was treated as a conduction-inspired empirical anchor. Its low-dimensional form was intended to remain informative when the learned correction had limited data support. Coefficient estimates and ablation results are reported in Section 4.3.

An RF residual model learned the departure from the greybox anchor. For each training observation, the residual target was

$$r_i=e_{001,i}^{meas}- e_{001,i}^{phys} \quad (13)$$

An RF regressor with 400 trees and a fixed random seed of 42 learned

$$\hat{r}_i = f_{RF}(\mathbf{x}_i) \tag{14}$$

Using the six-component vector in Equation 10 [37], the residual model represented dependencies absent from the greybox, including hatch spacing, focus offset, predicted width, and nonlinear interactions. These contributions were interpreted as model-based associations.

To limit residual correction under weak data support, the six inputs were standardized within each training fold and a kNN applicability weight was applied [26,27]:

$$\mathbf{z} = \frac{\mathbf{x} - \boldsymbol{\mu}_{train}}{\boldsymbol{\sigma}_{train}} \tag{15}$$

For a query $\mathbf{x}$, its applicability distance was defined as the mean Euclidean distance to the five nearest standardized training observations:

$$d(\mathbf{x}) = \frac{1}{5} \sum_{j \in N_5(\mathbf{x})} \|\mathbf{z} - \mathbf{z}_j\|_2 \tag{16}$$

Reference distances were computed leave-one-out with self matches excluded, so both query and training distances used five non-self-neighbors. The training-fold reference radius was the 95th percentile of these leave-one-out distances:

$$d_{ref} = P_{95}\left(\{d_i^{LOO}\}_{i \in train}\right) \tag{17}$$

The continuous residual weight was

$$w(x) = \exp\left[-\max\left(\frac{d(x)}{d_{ref}} - 1, 0\right)\right] \tag{18}$$

The weight equals unity when $d(x) \le d_{ref}$ and decreases smoothly beyond this radius. The final center prediction was

$$\hat{e}_{001}(\mathbf{x}) = e_{001}^{phys}(\mathbf{x}) + w(\mathbf{x})\, \hat{r}(\mathbf{x}) \tag{19}$$

Thus, the model retained the full residual correction for supported queries and reverted toward the greybox anchor as distance increased. The threshold $\tau_w$=0.60 fixed in the archived canonical configuration served only as a limited-support flag; it neither triggered abstention nor altered the conformal half-width. Four point predictors were compared under identical protocols: the greybox anchor $e_{001}^{phys}$; a 400-tree black-box random forest fitted directly to the measured target using the same six-component input vector; the ungated hybrid $e_{001}^{phys} + \hat{r}$; and the proposed gated hybrid $e e_{001}^{phys} + w \cdot \hat{r}$. Physics-domain abstention is defined separately in Section 3.4.

All data-dependent Stage-2 components were fitted within each training partition, as detailed in Section 3.5. After validation, the greybox, residual model, and applicability reference were refitted on the complete Stage-2 dataset for deployment.

**3.4 Group-CV conformal uncertainty quantification and physics-domain abstention**

Group-aware conformal intervals were constructed around the gated center prediction while preserving the experimental-group structure [28]. Because distribution shift weakens the standard exchangeability assumptions, coverage is reported as empirical retained-set coverage rather than as a strict conditional guarantee [30,31].

Within each outer evaluation split, the Stage-2 predictor was fitted on the outer-training observations, and inner LOGO generated out-of-group calibration predictions. Each inner validation group was predicted using a greybox, residual model, standardization, and applicability reference fitted without that group. The nonconformity score for held-out observation i was

$$s_i = |e_{001,i}^{meas} - \hat{e}_{001,i}^{(-g_i)}| \tag{20}$$

where $\hat{e}_{001,i}^{(-g_i)}$ denotes the gated prediction obtained without experimental group $g_i$. Only inner held-out observations satisfying $q_{PD,i} \leq \tau_q$ contributed nonconformity scores, with $\tau_q$=20,000 W mm$^{-2}$. This calibrate-on-issued rule restricted the calibration pool only. Observations above $\tau_q$ could therefore contribute to model fitting when they belonged to an inner-training group, but their nonconformity scores were excluded when they appeared in the inner held-out group. All Stage-2 components were refitted within each inner split, after which the complete predictor was fitted on the full outer-training set for prediction of the outer test fold.

For a nominal miscoverage level of $\alpha$=0.10, the retained calibration scores were sorted as $s_{(1)} \leq s_{(2)} \leq \cdots \leq s_{(n_{cal})}$. The conformal half-width was selected using the exact order statistic:

$$q_{0.90} = s_{(\lceil (n_{cal}+1)(1-\alpha) \rceil)} \tag{21}$$

All retained queries within the same outer evaluation split received the constant interval

$$C(\mathbf{x}) = \left[\hat{e}_{001}(\mathbf{x}) - q_{0.90}, \hat{e}_{001}(\mathbf{x}) + q_{0.90}\right] \tag{22}$$

The conformal half-width was not rescaled by $w(\mathbf{x})$ which affected only the center prediction through Section 3.3. A query with $w(\mathbf{x}) < \tau_w$ could therefore receive a physics-anchored prediction and interval when $q_{PD}(\mathbf{x}) \leq \tau_q$. The value $\tau_w$ alone did not trigger abstention. The black-box comparator used the same inner-group calibration structure and the same finite-sample order statistic. Its nonconformity scores were drawn from the same physics-valid inner held-out observations, so the two calibrations are directly comparable. The comparator itself implemented no abstention and issued a prediction and an interval for every test query.

A separate $q_{PD}$ criterion controlled the physics-validity decision. Using the focus-dependent beam radius from Equation 2, predictions and intervals were issued only when $q_{PD}(\mathbf{x}) \leq \tau_q$. Queries above this threshold were treated as outside the adopted deployment envelope of the conduction-inspired anchor, and both the center prediction and interval were withheld. The operational output was

$$O(\mathbf{x})=\begin{cases}\{\hat{e}_{001}(\mathbf{x}),C(\mathbf{x}),w(\mathbf{x})\}, & q_{PD}(\mathbf{x})\le\tau_q,\\ \text{abstain}, & q_{PD}(\mathbf{x})>\tau_q.\end{cases} \tag{23}$$

In the final canonical implementation, the threshold $\tau_q$ was fixed at 20,000 W mm$^{-2}$ and applied identically in every evaluation fold without re-estimation within folds. It lies above the largest value represented in the positive-defocus training domain ($1.10 \times 10^4$ W mm$^{-2}$) and below the smallest focus-zero value ($2.98 \times 10^4$ W mm$^{-2}$). Any threshold within this interval produces identical retention and abstention outcomes in all three evaluation settings. The available development record does not establish that the value was specified prospectively before inspection of the focus-zero results. It is a study-specific validity rule rather than a universal keyhole threshold. The weight $w(\mathbf{x})$ controls residual attenuation, whereas $\tau_q$ controls issuance. This separation prevents a data-support measure from being treated as an independent physics-validity guarantee. A query can therefore have full data support and still be withheld or lie far from the training distribution while remaining inside the validity envelope.

Coverage and interval width were evaluated only over retained predictions and reported together with retention. Abstained observations were reported separately and were not counted as uncovered predictions. This follows the selective-prediction principle while recognizing that the present rule is a study-specific regression-domain criterion rather than the original classification reject option [43]. Detailed metrics are defined in Section 3.5. Outer-test targets were excluded from model fitting, applicability-reference construction, and conformal calibration, as detailed in Section 3.5.

### 3.5 Validation protocol and statistical analysis

Stage-2 validation examined grouped generalization, controlled conduction-regime transfer, and behavior outside the adopted anchor-validity envelope. These settings represent distinct forms of model use. Stage-1 validation is described in Section 3.2 and summarized in Table 2.

The primary Stage-2 evaluation applied LOGO cross-validation to 278 observations from nine experimental groups. Each outer split withheld one complete group; all Stage-2 models, feature scaling, and applicability references were fitted using the remaining groups, with conformal calibration performed by inner LOGO as described in Section 3.4. The Stage-1 bridge was fitted once on the independent melt-pool dataset and frozen before Stage-2 partitioning. It used neither the EBSD-derived target nor the Stage-2 group labels and was not regenerated within the outer folds. LOGO therefore evaluates the Stage-2 mapping conditional on the frozen Stage-1 surrogate rather than end-to-end generalization.

The focus-offset holdouts were complementary physics-based stress tests rather than symmetric numerical extrapolations. At +80 mm, the increased surface beam radius reduced $q_{PD}$. The queries had weak residual-model support but remained within the adopted deployment envelope of the conduction-inspired anchor. At focus zero, the beam waist coincided with the powder-bed surface, producing the highest $q_{PD}$ values and exceeding $\tau_q$. Higher localized intensity is associated with the conduction-to-keyhole transition and vapor-depression formation [12,13], but $\tau_q$ is neither a universal keyhole criterion nor a basis for reclassifying the focus-zero observations as KM.

For the +80 mm evaluation, all 42 observations were withheld and the remaining 236 were used for fitting. Because this regime participated in framework development, it was treated as a controlled withheld-defocus evaluation rather than independent external validation. For the focus-zero evaluation, all 54 observations were withheld and the remaining 224 were used for fitting. Point-prediction metrics were calculated diagnostically, but no operational prediction or interval was issued because every focus-zero query exceeded $\tau_q$. Repeated process conditions were assigned entirely to either the training or withheld subset.

Within each evaluation regime, all Stage-2 models, feature scaling, and applicability quantities were recomputed from the training partition. No held-out target contributed to greybox fitting, residual learning, $d_{ref}$, or conformal calibration. Final metrics were regenerated after the canonical configuration was frozen; neighborhood-size and reference-percentile sweeps were treated only as robustness analyses.

Point-prediction performance was reported using $R^2$, mean absolute error (MAE), root-mean-square error (RMSE), and Spearman's ρ. LOGO metrics were calculated from pooled out-of-fold predictions and are therefore sample-weighted aggregates rather than unweighted means of group-specific scores. Withheld-defocus metrics were calculated over each complete withheld subset.

For evaluation regime g containing $n_g$ observations, let $R_g$ denote the retained set for which predictions and nominal 90% intervals $[L_i, U_i]$ were issued, and let $y_i$ denote the measured target. Retained-set coverage, mean full interval width, and retention were calculated as

$$\text{Coverage}_g = \frac{1}{|R_g|} \sum_{i \in R_g} \mathbf{1}\left( L_i \leq y_i \leq U_i \right) \tag{24}$$

$$\bar{W}_g = \frac{1}{|R_g|} \sum_{i \in R_g} \left( U_i - L_i \right) \tag{25}$$

$$\text{Retention}_g = \frac{|R_g|}{n_g} \tag{26}$$

Abstained observations remained in the retention denominator but were excluded from coverage and width because no interval was issued. Coverage was reported with retention and is not an unconditional guarantee across evaluation regimes. The non-abstaining black-box comparator was evaluated over all observations in each regime.

Interval width was compared with a regime-specific target-only null interval constructed from the corresponding retained set. The null interval was centered on the median target, with its half-width determined by applying the same finite-sample order-statistic rule to absolute deviations from that median. Relative width reduction was calculated as

$$\text{Width reduction}_g = 100 \left( 1 - \frac{\bar{W}_g}{W_{null,g}} \right) \% \tag{27}$$

Width reduction was reported only for the gated model. The target-only interval served as a descriptive width reference rather than a deployable baseline; coverage and width reduction were undefined when no predictions were retained.

Greybox-exponent stability was evaluated by resampling the nine experimental groups with replacement and refitting the log-linear model for 2,000 cluster-bootstrap replicates [44]. The Stage-1 predicted depth $\hat{D}$ was held in every replicate, so the percentile intervals were conditional on $\hat{D}$ and did not propagate Stage-1 uncertainty. Given nine groups, these intervals describe empirical group-resampling stability rather than complete workflow uncertainty or a causal scaling law. Variance inflation factors for logV, logP, and log$\hat{D}$ were computed to assess collinearity.

Residual-model interpretation used Shapley additive explanations (SHAP) and Sobol analysis at the controllable-process level [45–47]. P, V, H, and F were sampled independently from uniform distributions spanning their observed marginal bounds, while deterministic mediator surrogates supplied the frozen Stage-1 depth and width. Sobol first-order indices $S_1$, total-effect indices $S_T$, and the difference $S_T - S_1$ were reported, with the last used descriptively for interaction and higher-order contributions. The rectangular reference characterizes the fitted model under a synthetic independent-input distribution rather than the experimental design. TreeSHAP was applied to a deterministic attribution surrogate over the same reference space, and mean absolute contributions were normalized across the four inputs. The surrogate was not evaluated on an independent sample. Agreement between SHAP and Sobol therefore indicates consistency under a shared reference distribution, not independent validation. Pearson's (r) summarized the four-element profiles descriptively without inferential interpretation. Sampling and surrogate settings are provided in the Supplementary Information.

All calculations used fixed random seeds, and reported tables and figures were generated from saved per-observation and mechanism-analysis outputs.

# 4. Results and discussion

## 4.1 Grouped generalization performance

LOGO evaluation assessed generalization to the nine withheld experimental groups [25]. Each outer fold excluded one complete group, and the pooled out-of-fold predictions covered all 278 observations. Because process-variable values could overlap across groups, this protocol evaluates group-level generalization rather than formal covariate extrapolation.

The gated hybrid achieved $R^2$=0.592, compared with $R^2$=0.538 for the black-box RF model, 0.564 for the ungated hybrid, and $R^2$=0.287 for the greybox anchor (Table 3). The 0.054 gain over the black-box model and smaller 0.028 gain over the ungated hybrid indicated that residual learning provided most of the improvement over the anchor, with applicability weighting adding a smaller LOGO benefit. The gated model also reduced MAE and RMSE from 0.933 and 1.229 MUD for the black-box model to 0.856 and 1.156 MUD, while Spearman's r increased from 0.740 to 0.786.

**Table 3.** Point-prediction performance of the greybox anchor, black-box RF, ungated hybrid, and gated hybrid under LOGO and the two withheld-defocus evaluations.

| Model & evaluation regime | $R^2$ | MAE | RMSE | ρ |
|---|---|---|---|---|
| ***Physics-only (Greybox anchor)*** | | | | |
| LOGO generalization | 0.287 | 1.223 | 1.527 | 0.672 |
| Conduction (+80) | 0.778 | 0.558 | 0.680 | 0.886 |
| Out-of-envelope (0) | -0.298 | 1.480 | 1.883 | 0.581 |
| ***Black box*** | | | | |
| LOGO generalization | 0.538 | 0.933 | 1.229 | 0.740 |
| Conduction (+80) | -0.001 | 1.193 | 1.444 | 0.616 |
| Out-of-envelope (0) | -1.092 | 1.860 | 2.390 | 0.191 |
| ***Ungated hybrid*** | | | | |
| LOGO generalization | 0.564 | 0.891 | 1.194 | 0.773 |
| Conduction (+80) | 0.199 | 1.096 | 1.291 | 0.677 |
| Out-of-envelope (0) | -0.770 | 1.732 | 2.199 | 0.256 |
| ***Gated Hybrid (This work)*** | | | | |
| LOGO generalization | 0.592 | 0.856 | 1.156 | 0.786 |
| Conduction (+80) | 0.750 | 0.588 | 0.722 | 0.847 |
| Out-of-envelope (0) | -0.378 | 1.536 | 1.940 | 0.431 |

These aggregate scores do not establish how the models behave when an entire process sub-regime is withheld. Figure 4 positions the LOGO, +80 mm, and focus-zero evaluations in the applicability-validity plane, motivating the complementary regime-shift comparisons in Section 4.2.

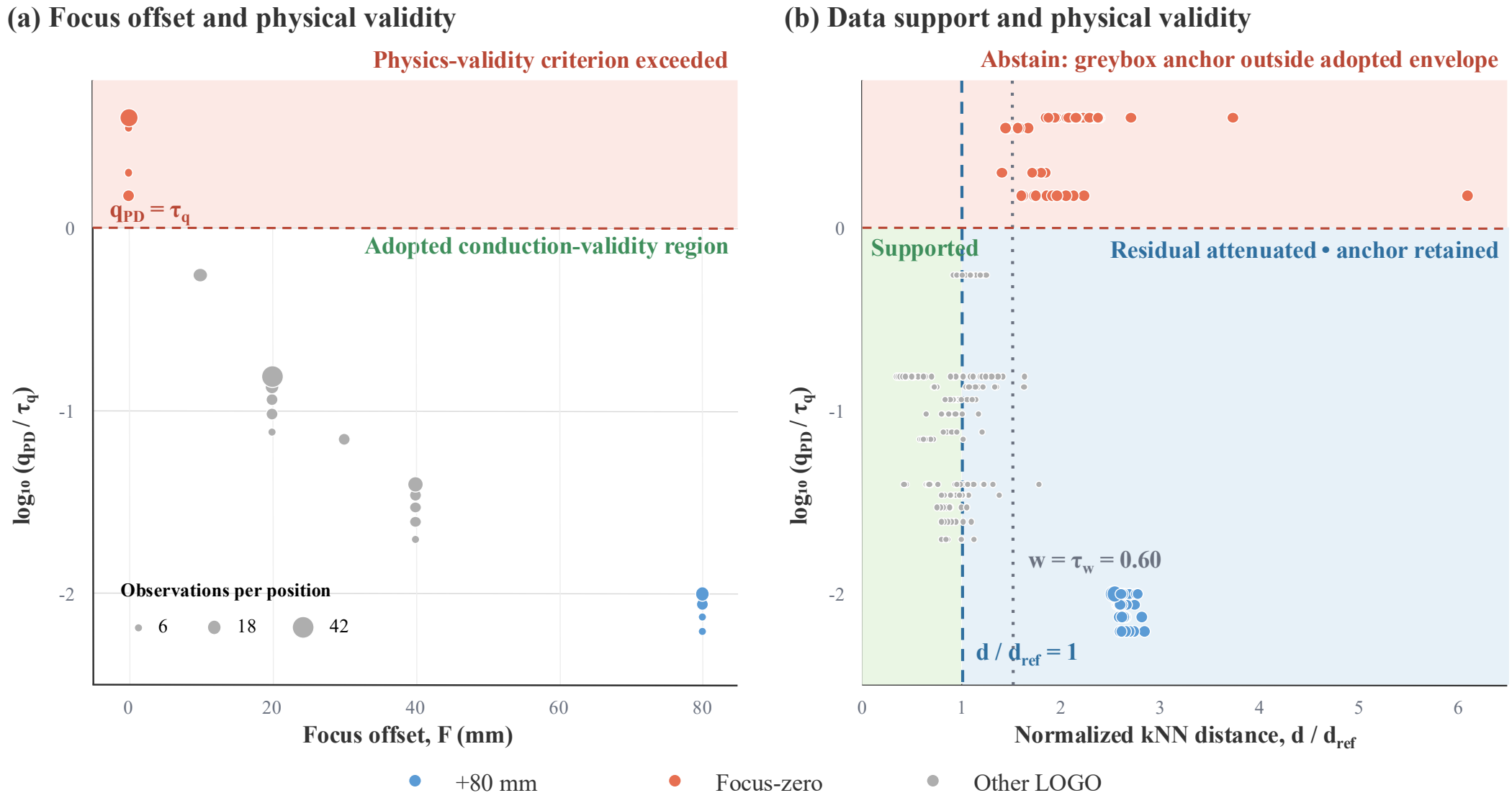

**Figure 4.** Evaluation design in the applicability-validity plane. (a) $\log_{10}\left(q_{PD}/\tau_q\right)$ versus focus offset. Marker area represents the number of coincident observations at each of the 20 unique (P, F) positions. (b) The same validity coordinate versus the evaluation-specific normalized five-nearest-neighbor distance $d/d_{ref}$. The horizontal dashed line marks $q_{PD}=\tau_q$, above which predictions are withheld. The blue dashed line marks $d/d_{ref}=1$ where residual attenuation begins, and the gray dotted line marks $w=\tau_w=0.60$ corresponding to $d/d_{ref}\approx1.511$. Colors identify the complete +80 mm holdout (n=42), complete focus-zero holdout (n=54), and remaining LOGO observations (n=182). Because $d/d_{ref}$ depends on the training partition, panel (b) uses values from the corresponding LOGO, +80 mm, or focus-zero evaluation; $q_{PD}/\tau_q$ is partition-independent.

### 4.2 Reliability under withheld-defocus regime shift

Withholding the complete +80 mm regime (n=42) produced a clearer separation among model classes. All 42 queries were routed to CM by the frozen Stage-1 surrogate. Figure 5 compares parity under LOGO and +80 mm evaluation. The latter probes transfer under weak residual-model support but a retained greybox anchor, complementing the focus-zero test of abstention beyond the adopted physics-validity criterion.

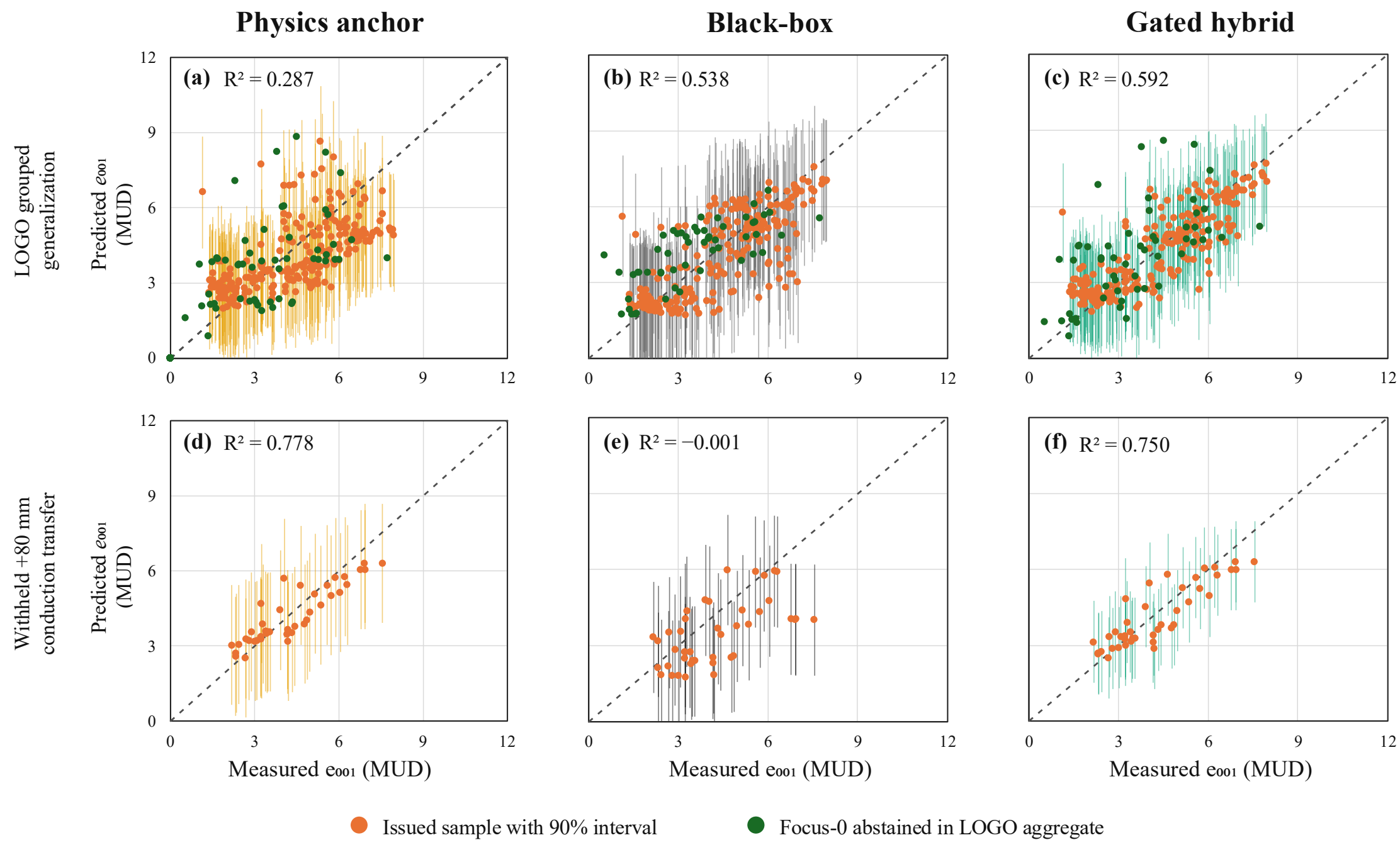

**Figure 5.** Regime-specific parity of measured and predicted $e_{001}$ for the greybox anchor, black-box RF, and gated hybrid. Panels (a-c) show LOGO results; panels (d–f) show the withheld +80 mm evaluation. Dashed lines denote 1:1 agreement. Orange markers identify framework-retained observations, whereas green markers identify focus-zero observations for which the framework abstained. Vertical bars show model-specific 90% inner group-CV empirical intervals where issued. The black-box comparator does not implement abstention; its full-set UQ is reported in Table 7. All axes span 0-12 MUD.

At +80 mm, the greybox anchor achieved $R^2$=0.778, whereas the black-box RF reached $R^2$= -0.001 and the ungated hybrid reached $R^2$=0.199. The near-zero black-box result was equivalent to the withheld-subset mean baseline under $R^2$, while unrestricted residual correction did not preserve the transfer behavior of the anchor. The gated hybrid achieved $R^2$=0.750, remaining close to the greybox and substantially outperforming the black-box and ungated models, although it did not improve on the anchor. This pattern is consistent with attenuation of an unsupported residual correction. The anchor component was therefore associated with the observed transfer, and Section 4.3 examines the contribution of its Stage-1 depth mediator.

Focus-zero defined a different boundary. This subset contained 54 observations at the highest

$q_{PD}$ values, and the greybox, black-box RF, and gated hybrid produced negative $R^2$ values of -0.298, -1.092, and -0.378 respectively. Stage-1 nevertheless routed 53 observations to CM and one to KM, so the subset is described as outside the adopted anchor-validity envelope rather than as keyhole or transitional regime [9,12–14]. The mode label and Stage-2 validity decision therefore represent distinct assessments. Because all 54 queries exceeded $\tau_q$, the framework withheld every operational prediction and interval, giving 0% retention. Focus-zero consequently serves as a failure-case stress test rather than evidence of successful extrapolation.

Because the +80 mm regime contributed to framework development and configuration assessment, it was treated as a controlled withheld-defocus evaluation rather than as independent external validation. Within this entirely CM-routed subset, measured $e_{001}$ reached 7.557 MUD, showing that pronounced enrichment was achievable within the evaluated conduction-oriented domain. However, with only one KM-routed observation among the 278 Stage-2 cases and no matched porosity or defect labels, the data do not support a comparison of defect-free CM texture with defect-associated KM texture.

Reported accuracies from LPBF texture-prediction studies cannot be interpreted as like-for-like rankings because the studies differ in material, target definition, model output, and validation design. Sofras et al. [21] used decision-tree regression, with ten-fold cross-validation for tree pruning and evaluated six newly fabricated control conditions to predict neutron-diffraction-derived texture ratios in 304L stainless steel. Whitney et al. [22] combined part-scale mechanistic simulation with a ML surrogate for Ti-6Al-4V microstructure prediction, and its reported surrogate accuracy concerns phase-fraction and lath-width outputs rather than texture. Table S5 in the Supplementary Information compares the reported targets, metrics, computational pathways, and applicability treatments without treating them as directly ranked accuracies.

Quantitative model comparison in the present study was instead performed on the same dataset under identical evaluation protocols (Table 3). Under LOGO, the proposed gated hybrid achieved $R^2$ = 0.592, compared with 0.538 for the black-box model and 0.287 for the physics anchor. In the +80 mm evaluation, it retained most of the anchor performance (0.750 versus 0.778), whereas the black-box model reached −0.001. Corresponding mean absolute percentage errors are reported in Table S6. The reimplemented and adapted baselines in Table S7 remained below the physics anchor at +80 mm across all predictor sets and across three pruning implementations. Neither representative study in Table S5 reports the query-level combination of residual attenuation and validity-based output withholding implemented here.

As a post hoc diagnostic comparison, literature-derived model forms were reimplemented or adapted to the present dataset and evaluated under identical protocols (Table S7). A single regression tree with cost-complexity pruning, following the procedure of Sofras et al.[21], achieved ($R^2$=0.191) under LOGO and 0.175 on the +80 mm subset using the published predictors of laser power, scan speed, and hatch spacing. Adding focus offset as a dataset-specific adaptation increased these values to 0.302 and 0.297, respectively. Power-law regressions using conventional volumetric and linear energy-density descriptors [48] provided no positive predictive skill under either LOGO or the +80 mm holdout. These descriptors do not account for focus offset; for example, four conditions at 400 W, 600 mm $s^{-1}$, and 150 μm shared a volumetric energy density of 148 J $mm^{-3}$ despite measured ($e_{001}$) values spanning 2.42–7.84 MUD. A normalized-enthalpy descriptor following the dimensionless formulation used for LPBF melting-mode analysis [12], with the adopted IN718 properties and focus-dependent beam radius, achieved ($R^2$=0.028) at +80 mm, whereas a reduced geometric surrogate motivated by the melt-pool and thermal-gradient growth framework of Liu et al. [49] achieved 0.013. Across all predictor sets and across three pruning implementations, these baselines remained below the physics anchor at +80 mm.

Together, the holdouts distinguish anchor-supported transfer within the adopted envelope from abstention beyond it. Sections 4.4 and 4.5 examine the corresponding residual-attenuation and physics-validity decisions.

### 4.3 Greybox scaling relation and depth-mediator ablation

Log-linear fitting across the 278 observations yielded the coefficients of the P-V-$\hat{D}$ greybox anchor defined in Equation. 12 (Table 4). The positive exponents for V and P together with the negative exponent for $\hat{D}$ represent conditional log–log associations within the fitted process envelope. All three 95% experimental-group cluster-bootstrap intervals excluded zero, and the variance inflation factors of the log-transformed predictors were below 2.5 (Table 5). These results indicate stable coefficient signs without severe linear collinearity, but they do not establish universal physical constants or causal effects.

**Table 4.** Fitted coefficients of the P-V-$\hat{D}$ greybox anchor and 95% experimental-group cluster-bootstrap percentile intervals (2,000 replicates).

| Term | Point estimate | Bootstrap median | 95% CI | Expected sign | Interpretation |
|---|---|---|---|---|---|
| V | 0.296 | 0.288 | [0.024, 0.452] | + | higher speed associated with stronger $e_{001}$ |
| P | 0.407 | 0.412 | [0.271, 0.906] | + | higher power associated with stronger $e_{001}$ |
| $\hat{D}$ | -0.629 | -0.592 | [−0.798, −0.318] | − | deeper melt pool associated with weaker $e_{001}$ |
| Prefactor A | 0.0148 | — | — | + | $e_{001}^{Phys} = 0.0148\ V^{0.30} P^{0.41}\ \hat{D}^{-0.63}$ |

*Bootstrap intervals quantify coefficient stability across group resamples and are not interpreted as universal physical constants.*

**Table 5.** Collinearity diagnostics and leakage-free greybox ablation under LOGO and the withheld +80 mm evaluation.

| Specification / diagnostic | VIF / LOGO $R^2$ | +80 mm $R^2$ | Outcome |
|---|---|---|---|
| VIF (log V) | 1.48 | — | below 2.5 |
| VIF (log P) | 1.41 | — | below 2.5 |
| VIF (log $\hat{D}$) | 1.06 | — | below 2.5 |
| {V, P, $\hat{D}$} selected | 0.287 | 0.778 | retained anchor |
| + $E_V$ (volumetric energy) | 0.314 | 0.747 | small LOGO gain; reduced +80 transfer |
| + H | 0.314 | 0.747 | same column-space as $E_V$; reduced +80 transfer |
| + \|F\| | 0.245 | 0.699 | transfer weakened |
| V, P only (drop $\hat{D}$) | 0.059 | -0.116 | transfer failed |
| P, D only (drop V) | 0.141 | 0.317 | transfer substantially degraded |

*Variance inflation factors were calculated for the log-transformed predictors. Each candidate specification was refitted within the corresponding training partition.*

Leakage-free ablation identified $\hat{D}$ as central to the transfer behavior of the anchor. Removing $\hat{D}$ reduced LOGO $R^2$ from 0.287 to 0.059 and the withheld +80 mm $R^2$ from 0.778 to −0.116 (Table 5). The model-derived depth therefore carried predictive information under the withheld-defocus evaluation, although this result establishes predictive relevance rather than a causal effect of melt-pool depth.

Adding $E_V$ or a single multiplicative hatch term produced identical scores because, for fixed layer thickness t, $\log E_V = \log P - \log V - \log H - \log t$. Once P and V are included, these additions span the same column space [15]. Both slightly improved LOGO performance but reduced +80 mm transfer. Adding |F| or removing V also weakened transfer, favoring the parsimonious P-V-$\hat{D}$ specification. The result for the single hatch term does not exclude nonlinear or interaction-related hatch effects, which are examined through the learned residual in Section 4.6.

Together, the coefficient and ablation results suggest that $\hat{D}$ compresses part of the process-to-melt-pool response into a geometry mediator, while the multiplicative form constrains the anchor to a small set of variables. This structure retained predictive value at +80 mm but remained incomplete under LOGO, where the anchor achieved $R^2$=0.287. This ablation establishes the predictive relevance of the frozen Stage-1 mediator within the present workflow; it does not establish a causal effect of melt-pool depth or a universal physical scaling law. Fourteen of the 278 Stage-2 observations lay outside the Stage-1 hatch-spacing range. For these cases, the frozen tree models generated the geometry mediators beyond the observed hatch-spacing range; these outputs should not be interpreted as interpolated estimates. Section 4.4 next tests whether applicability weighting can suppress unsupported residual corrections without sacrificing anchor-supported transfer.

### 4.4 Applicability weighting and gate ablation

Because unrestricted residual correction improved LOGO performance but degraded +80 mm transfer, the distance-based applicability weighting defined in Equations 15–19 [26] was evaluated through a leakage-free comparison of three gate configurations. The deployed soft gate achieved $R^2$=0.592 under LOGO and 0.750 at +80 mm, compared with 0.564 and 0.199 for the ungated hybrid (Table 6, Figure 6a). For the +80 mm queries, the soft gate produced a mean residual attenuation of 80.5%, corresponding to a mean residual weight of approximately 0.195. The hard gate assigned zero residual weight to all +80 mm queries and recovered the standalone anchor result of $R^2$=0.778, but its LOGO performance decreased to 0.530. Among the evaluated gates, continuous attenuation therefore provided the most balanced trade-off between grouped generalization and anchor-supported transfer.

**Table 6.** Gate-type ablation under LOGO grouped generalization and the withheld +80 mm evaluation.

| Gate configuration | LOGO $R^2$ | +80 mm $R^2$ | Mean +80 mm attenuation |
|---|---|---|---|
| Soft gate (deployed) | 0.592 | 0.750 | 0.805 |
| Ungated hybrid (plain) | 0.564 | 0.199 | 0.000 |
| Hard cutoff | 0.530 | 0.778 | 1.000 |

*Residual attenuation is the +80 mm mean of (1-w). The ungated, hard, and soft gates apply full, binary, and exponentially attenuated residual corrections, respectively.*

The deployed k=5, $P_{95}$ configuration was inherited from the frozen canonical implementation and was not selected from the sensitivity sweeps. Although k=3 produced a slightly higher +80 mm $R^2$, the qualitative gate trade-off remained unchanged across the evaluated neighborhood sizes. Larger neighborhoods and the more permissive $P_{99}$ reference radius allowed greater residual correction and reduced +80 mm transfer (Figure 6b, c). These leakage-free refits were used only to assess robustness around the deployed configuration; complete sweep results are provided in the Supplementary Information.

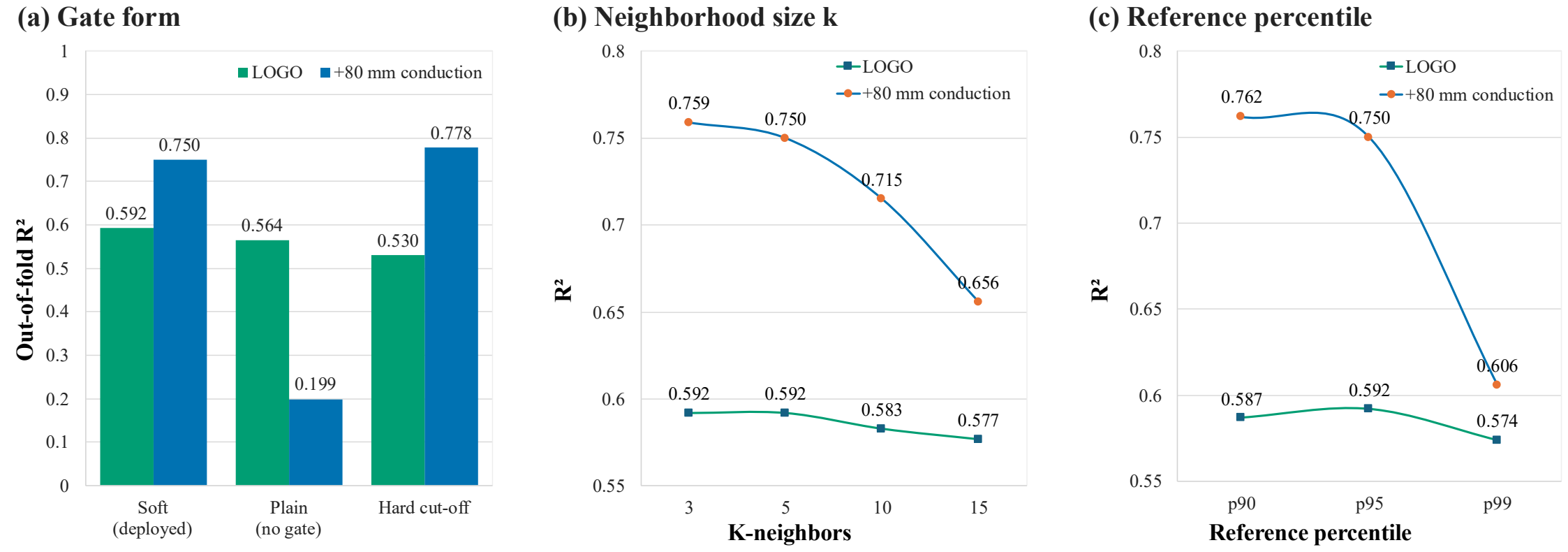

**Figure 6.** Applicability-gate ablation and sensitivity analysis. (a) LOGO grouped generalization and withheld +80 mm $R^2$ for the ungated hybrid, hard gate, and deployed soft gate. (b) Sensitivity of the soft gate to the neighborhood size k. (c) Sensitivity to the reference-distance percentile. The deployed $k=5$, $P_{95}$ configuration was not selected from these leakage-free robustness sweeps.

The gate ablation shows that applicability weighting controls the magnitude of the learned residual without altering the greybox anchor. The soft gate retained the highest LOGO performance while keeping the +80 mm response close to the standalone anchor. However, the distance weight is neither an uncertainty guarantee nor an independent abstention rule. Section 4.5 evaluates retained-set interval coverage and the separate physics-validity criterion used to withhold unsupported predictions.

### 4.5 Group-CV conformal uncertainty and physics-validity abstention

Reliability was evaluated through nominal 90% group-CV conformal intervals for retained predictions and a separate physics-validity abstention rule. The continuous applicability weight $w(\mathbf{x})$ attenuated the learned residual, while $w<\tau_w$ identified limited data support without changing interval width or independently triggering abstention. Predictions were withheld only when $q_{PD}>\tau_q$, indicating that the greybox anchor was outside its adopted validity envelope. Coverage is therefore reported as empirical retained-set coverage together with retention, rather than as unconditional coverage under distribution shift.

Under LOGO, 224 of the 278 observations were retained, giving 80.6% retention. The gated intervals achieved 92.9% empirical retained-set coverage with a mean full width of 3.65 MUD, 34% narrower than the regime-specific null width of 5.57 MUD (Table 7). The black-box intervals covered 91.0% with a mean width of 4.76 MUD but were evaluated over all 278 observations because the comparator did not abstain. The LOGO coverage values therefore refer to different evaluation populations and should not be compared without considering retention.

**Table 7.** Empirical interval coverage, mean full width, and retention under LOGO and the two withheld-defocus evaluations.

| Regime | Gated empirical retained-set coverage | Gated width (MUD) | Gated retention | Black-box coverage | Black-box width |
|---|---|---|---|---|---|
| LOGO generalization | 92.9% | 3.65 | 80.6% | 91.0% | 4.76 |
| +80 mm | 100% | 3.21 | 100% | 83.3% | 4.39 |
| Focus-zero | — | — | 0% | 53.7% | 3.32 |

*Gated coverage and width are calculated only for issued intervals; black-box values use the complete evaluation subset because the comparator does not abstain. The dash indicates that no gated interval metric is defined at zero retention.*

At +80 mm, all 42 observations satisfied the physics-validity criterion and were retained, although every query had $w<\tau_w$, indicating limited residual-model support. The gated intervals achieved 100% empirical coverage with a mean full width of 3.21 MUD, 40% narrower than the regime-specific null width of 5.34 MUD. Because retention was 100%, the gated and black-box intervals were evaluated over the same observations; the black-box comparator achieved 83.3% coverage with a mean width of 4.39 MUD.

At focus-zero, all 54 observations exceeded $\tau_q$, so no gated predictions or intervals were issued and retention was 0%. Gated coverage and width are consequently undefined. Without abstention, the black-box intervals had a mean width of 3.32 MUD but covered only 53.7% of the observations. Thus, comparatively narrow intervals did not indicate reliable prediction outside the adopted physics-validity envelope (Figure 7).

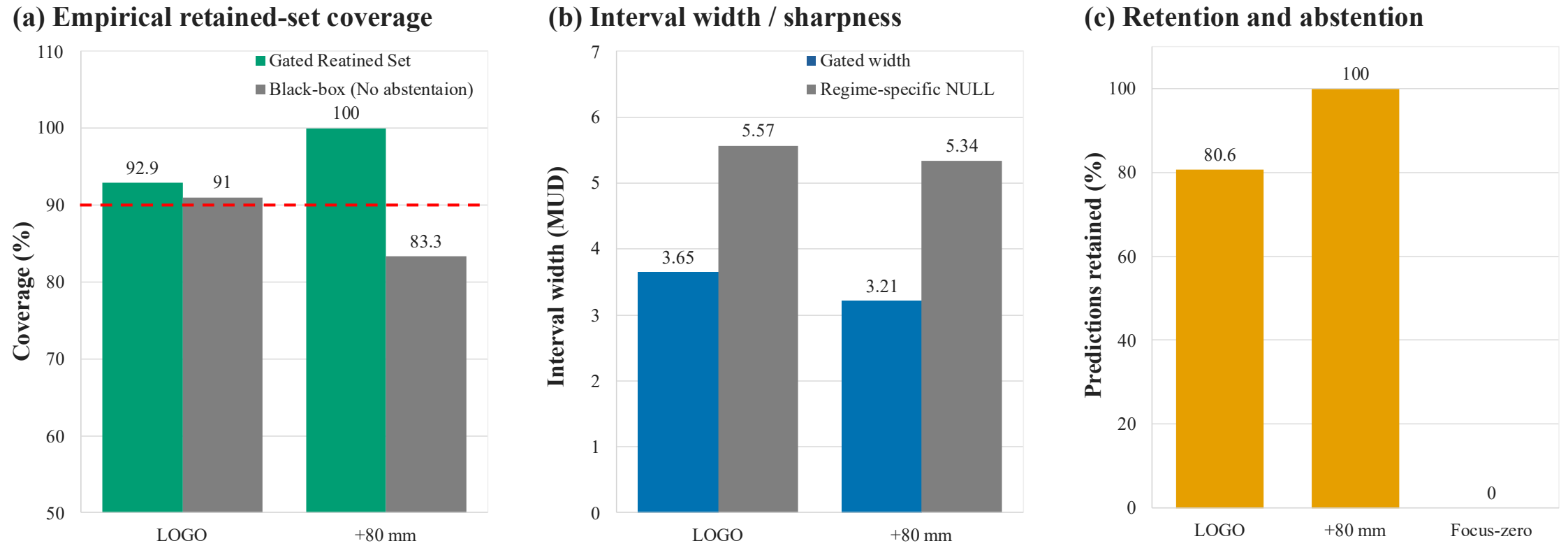

**Figure 7.** Empirical uncertainty and selective prediction across the evaluation regimes. (a) Coverage of issued gated intervals and full-set black-box intervals; no gated interval was issued at focus-zero. (b) Gated mean full interval width and the regime-specific target-only null width for evaluations with nonempty retained sets. (c) Gated retention. The horizontal reference in panel (a) denotes the nominal 90% coverage level.

For both nonempty retained sets, empirical coverage was numerically above the nominal 90% level while the intervals remained narrower than the corresponding target-only null widths. However, the 100% coverage at +80 mm was obtained from 42 observations under a specific regime shift and is not a theoretical guarantee under arbitrary distribution shifts or violations of exchangeability [30]. As noted in Section 3.4, the physics-validity threshold is a study-specific rule rather than a prospectively specified constant. Coverage and retention together support an operational assessment within the evaluated protocols. Section 4.6 examines the process variables and interactions represented by the learned residual.

### 4.6 Process-level interpretation of the learned residual

To interpret the fitted residual at the process level, SHAP and Sobol analyses were applied to the complete process-to-residual mapping, with predicted melt-pool depth and width supplied by Stage-1 mediator surrogates [45–47].

Normalized mean absolute SHAP attributed 37% of the residual contribution to scan speed (V),

31% to hatch spacing (H), 27% to focus offset (F), and 4% to power (P) (Figure 8a; rounded values sum to 99%). Scan speed remained the largest contributor despite its explicit inclusion in the greybox, which is consistent with speed-dependent nonlinearities or interactions not represented by the single fitted exponent $V^{0.30}$, rather than with an omitted process variable.

Under the same uniform process-variable bounds, Sobol analysis ranked hatch spacing $S_1$=0.338 and $S_T$=0.449, and scan speed $S_1$=0.311 and $S_T$=0.416 as the leading variables (Figure 8b). Focus offset had $S_1$=0.183 and $S_T$=0.286, whereas power had a limited total effect of 0.040. Its small negative first-order estimate of −0.002 was treated as numerical variation around zero. The reversal of the leading SHAP and Sobol rankings is expected because mean absolute SHAP summarizes attribution magnitude, whereas Sobol indices allocate output variance under the specified reference distribution.

The interaction indicator $S_T$-$S_1$ was 0.111 for hatch spacing, 0.105 for scan speed, 0.103 for focus offset, and 0.041 for power (Figure 8a, b). Hatch spacing was therefore the leading variable absent from the greybox under both residual-SHAP attribution and the Sobol interaction profile. This result resolves the apparent tension with Section 4.3: a single global log-linear hatch term did not improve the transfer-oriented greybox, whereas the learned residual represented nonlinear and interaction-dependent hatch behavior. Such behavior is consistent with the effects of hatch spacing and remelting on track overlap, repeated thermal exposure, and competitive grain selection [8,10,11], although the attribution does not identify a specific physical pathway. Because $\widehat{D}$ and $\widehat{W}$ entered through model-derived mediators, these indices describe the complete residual forward mapping and do not uniquely separate direct process contributions from those transmitted through Stage 1.

Across the four process variables, the residual-SHAP and Sobol-interaction profiles showed a descriptive magnitude correspondence of (r=0.96) (Figure 8c). Because both analyses used the same synthetic reference bounds and the comparison contained only four values, this correlation is an internal consistency summary rather than inferential evidence, independent validation, or a causal result.

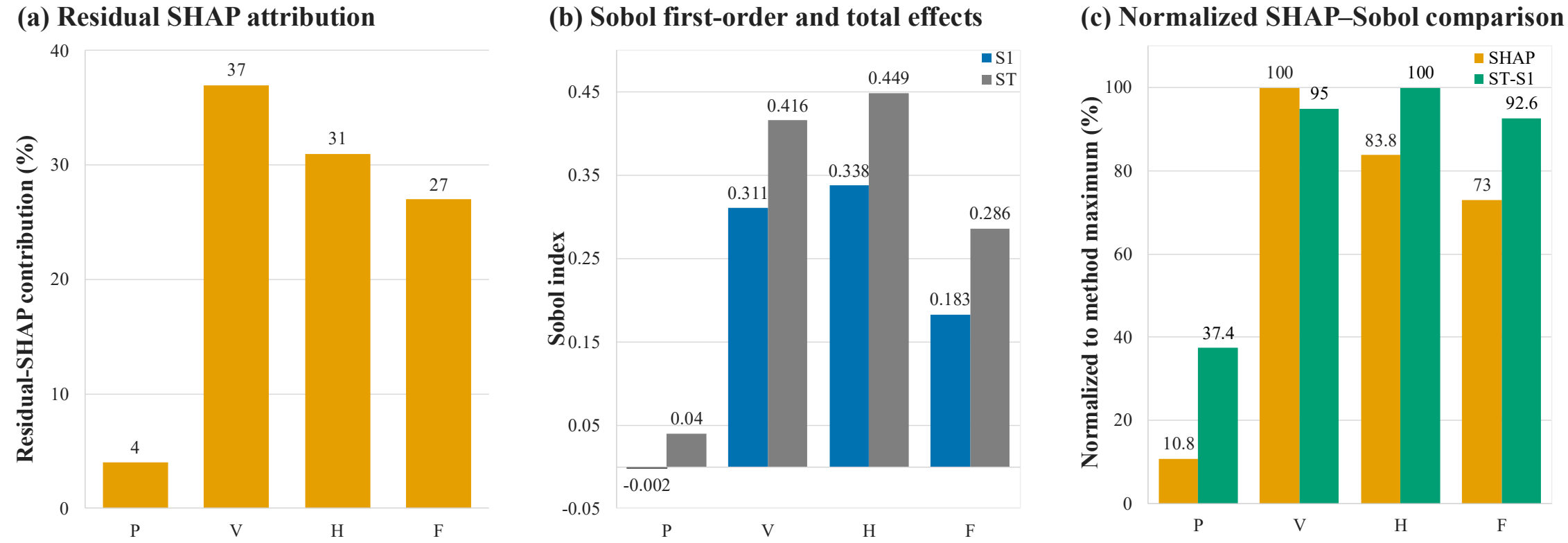

**Figure 8.** Process-level interpretation of the fitted residual forward mapping, with Stage-1 melt pool depth and width treated as model-derived mediators. (a) Normalized mean absolute SHAP attributions. (b) Sobol first-order ($S_1$) and total-effect ($S_T$) indices under the shared uniform process-variable bounds. (c) Independently normalized comparison of the SHAP attribution and Sobol interaction term ($S_T$-$S_1$). Pearson's (r=0.96) was calculated from the unnormalized four-variable profiles and is reported descriptively.

Together, the two analyses associate the learned residual with nonlinear speed dependence and hatch- and focus-related interactions omitted from the parsimonious greybox. Because this structure was learned from the sampled process space, applicability weighting remains necessary when the residual model is queried under limited data support. Hatch spacing was the leading process variable absent from the anchor, but this does not conflict with the ablation in Section 4.3. A single multiplicative hatch term was log-linearly equivalent to the tested energy-density extension, whereas the residual learner could represent nonlinear and interaction-dependent hatch behavior. Because $\hat{D}$ and $\hat{W}$ entered as Stage-1 mediators, the process-level sensitivities do not uniquely separate direct from mediated contributions. Section 5 applies the retained texture predictions in an illustrative literature-anchored mapping to build-direction elastic modulus.

## 5. Illustrative application to build-direction elastic modulus

This section presents two complementary analyses that connect ⟨001⟩ ‖ BD enrichment to property-scale behavior. Section 5.1 applies a bounded, literature-informed texture-to-modulus mapping to the retained predictions and a separately built set. Section 5.2 uses tensile responses from the same specimens to examine whether the assumed inverse association between enrichment and directional stiffness persists at room temperature and at 650 °C. This experimental comparison supports the direction of the mapping, not its numerical calibration.

### 5.1 Texture-to-modulus mapping and application

Build-direction Young's modulus, $E_{BD}$, was selected to provide a property-scale interpretation of changes in ⟨001⟩ ‖ BD enrichment. Elastic stiffness in cubic nickel-based alloys is determined by orientation through the single-crystal stiffness tensor [6,7], so texture is a first-order control on modulus. Strength is instead governed primarily by the γ″ and γ′ precipitation state [50], with grain size and porosity as further contributors, and is therefore not mapped here; a direct check of this separation on the present specimens is reported in Section 5.2.

Directional stiffness is a design-relevant quantity in components subject to constrained thermal cycling, where the stress developed under a given thermal strain scales with the modulus along the constrained direction. LPBF IN718 is used for hot-section components in aero-engines and turbomachinery [51], and directional stiffness has been exploited at the design stage to reduce machining-induced deformation in thin-walled LPBF parts[3], and because epitaxial growth follows the build direction [52], the build orientation chosen at the design stage fixes the texture the part will carry. Whether a lower build-direction modulus is desirable is application dependent and is not claimed here.

$E_{BD}$ was defined as a monotone nonincreasing function of the predicted texture intensity. [6,7]:

$$E_{BD} = \text{clip}\left[195 - 10.2\,(\hat{e}_{001} - 1),\ 125,\ 195\right] \tag{28}$$

where $\hat{e}_{001}$ is the predicted ⟨001⟩‖BD texture intensity. A slope of 10.2 GPa per unit $\hat{e}_{001}$ was adopted, which places the lower saturation bound at $\hat{e}_{001} \approx 7.9$ close to the maximum value observed in the main dataset. This is of the same order as the gradient implied by reported build-direction moduli for LPBF IN718 at two texture levels, which differ by 25 GPa across a change in ⟨001⟩ pole-figure intensity of about 2.8 [3]. The two texture measures are defined differently, and the comparison is indicative only. The clip operator restricts $E_{BD}$ to 125–195 GPa.

Let g denote the clipped mapping in Equation 28. Because g is monotone nonincreasing, a texture interval $[L_e, U_e]$ propagates to

$$[L_E, U_E] = [g(U_e), g(L_e)] \tag{29}$$

The bounded, literature-informed relation in Equation 28 was applied to the 224 retained LOGO texture predictions to illustrate downstream property interpretation. For cubic IN718, stronger predicted ⟨001⟩ ∥ BD texture maps to a lower build-direction elastic modulus $E_{BD}$ consistent with the expected elastic anisotropy [3–8,53]. The mapped values ranged from 126.9 to 186.6 GPa, reported as approximately 127-187 GPa in Figure 9. Only retained texture predictions and their propagated intervals were mapped; no modulus was reported for the abstained focus-zero observations.

The 90% texture intervals were propagated through the same monotonic relation using Equation 29. The mean pre-clipping modulus half-span across the retained set was 18.6 GPa, although it varied among observations because the texture intervals were fold-dependent. Clipping at the prescribed 125 and 195 GPa bounds also made some propagated intervals asymmetric. These intervals represent texture-prediction uncertainty propagated through the assumed mapping; they do not include uncertainty in the mapping form or material constants. This transformation propagates retained texture predictions and intervals to a property-oriented quantity while preserving abstention.

A separate set of specimens was subsequently built and machined, and their ⟨001⟩ ∥ BD enrichment was measured on the tensile bars themselves. Neither their process settings nor their measurements entered model development. The framework was applied to these nine conditions with all fitted parameters and both gate thresholds held at their canonical values. The framework separated these nine conditions into three groups (Table 8). Three exceeded the adopted anchor-validity threshold by a factor of four and were withheld without further prediction or interval. Three lay at normalized applicability distances near 3.1, where the learned correction was attenuated to about 12% of its unweighted value, leaving predictions within 0.07 MUD of the physics anchor alone. The remaining three were issued with a mean absolute error of 1.02 MUD, or 8 to 17% of the measured value. Mapped through Equation 28, these three gave build-direction moduli within 8.7 GPa on average of the values implied by the measured enrichments, and all three propagated intervals contained the measured value (Figure 9). Across the six conditions for which predictions were issued or attenuated, the predicted and measured enrichments ranked identically. Two of the nine conditions also appear in the main dataset and gave enrichments of 8.38 and 6.47 against 7.84 and 2.42 measured previously on separately built specimens, indicating that nominally identical process settings need not reproduce the same texture level across builds.

**Table 8.** Framework decisions on nine conditions from a separately built set. Values in bold exceed the corresponding threshold: the anchor-validity criterion withholds a prediction when $q_{PD}$ exceeds $\tau_q$, and the applicability weighting attenuates the learned correction when the normalized distance exceeds the value corresponding to $\tau_w$. No prediction was issued for the withheld conditions. For the attenuated conditions the correction was reduced to about 12% of its unweighted value, so the reported values are close to the physics anchor alone. Measured enrichments were obtained on specimens that took no part in model development.

| Condition (P-S-H-F) | $q_{PD}/\tau_q$ | $d/d_{ref}$ | w | Decision | Measured $e_{001}$ | Predicted $e_{001}$ |
|---|---|---|---|---|---|---|
| 400-400-150-0 | 3.980 | 0.99 | 1.00 | withheld | 3.49 | N/A |
| 400-400-250-0 | 3.980 | 0.87 | 1.00 | withheld | 2.99 | N/A |
| 400-600-150-0 | 3.980 | 0.35 | 1.00 | withheld | 6.47 | N/A |
| 400-400-150-(-80) | 0.010 | 3.11 | 0.12 | attenuated | 4.92 | 3.79 |
| 400-400-250-(-80) | 0.010 | 3.11 | 0.12 | attenuated | 1.15 | 3.67 |
| 400-600-150-(-80) | 0.010 | 3.04 | 0.12 | attenuated | 7.04 | 4.20 |
| 400-400-150-(+20) | 0.153 | 0.50 | 1.00 | issued | 7.43 | 6.15 |
| 400-400-250-(+20) | 0.153 | 0.56 | 1.00 | issued | 7.07 | 5.99 |
| 400-600-150-(+20) | 0.153 | 0.24 | 1.00 | issued | 8.38 | 7.69 |

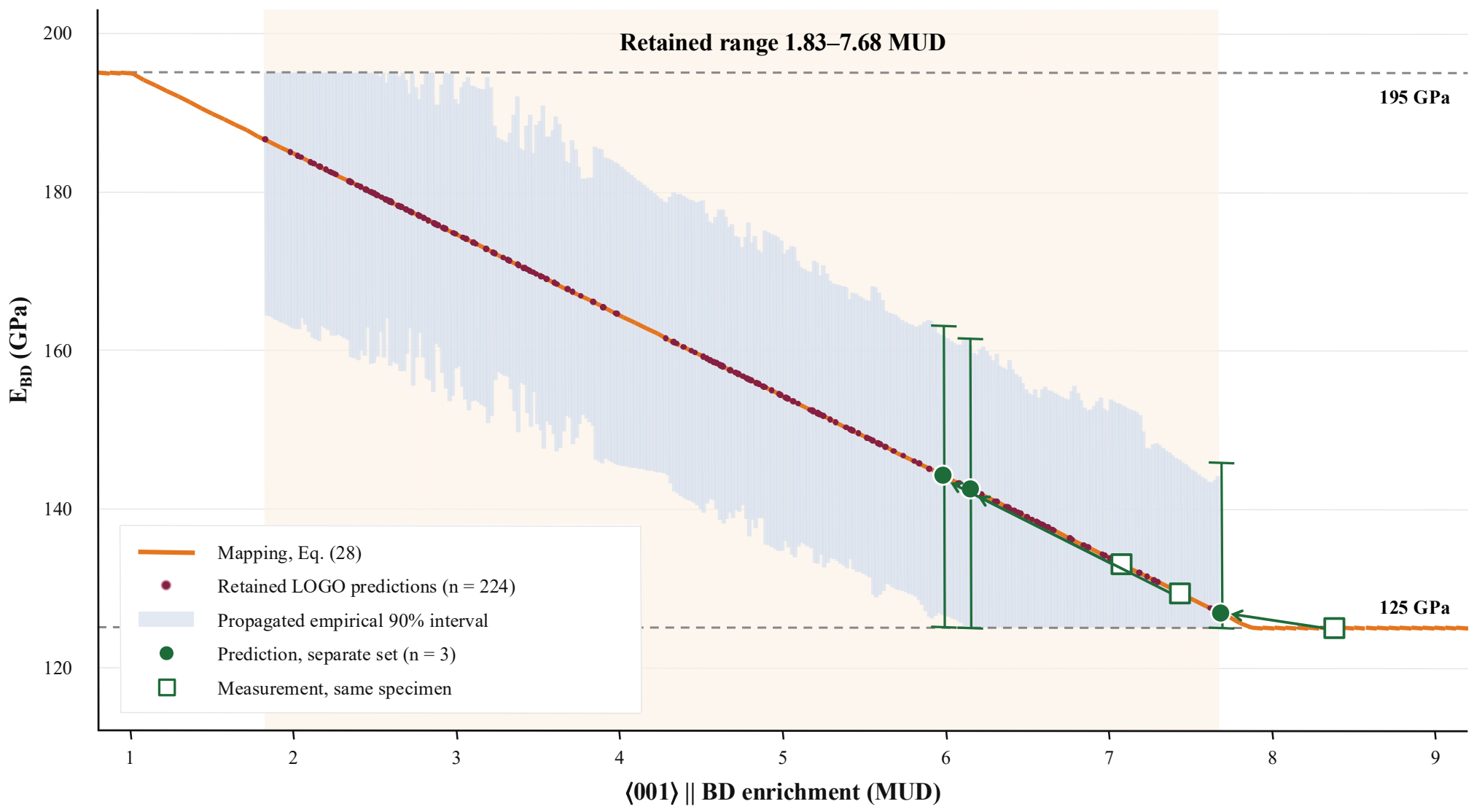

**Figure 9.** Illustrative mapping from ⟨001⟩ ‖ BD enrichment to build-direction elastic modulus, and its application to a separately built set. The orange background marks the retained texture range and the blue band the propagated empirical 90% intervals of the 224 retained LOGO predictions. For the separately built set, filled circles are the three predictions that satisfied both gate criteria, open squares are the moduli obtained by applying the same mapping to the enrichment measured on the corresponding specimens, and arrows connect each pair. Their error bars were propagated from the median retained-LOGO texture half-width; all three contained the measured value. The withheld and attenuated conditions are not shown.ss

## 5.2 Directional response at service temperature

Because the mapping in Equation 28 rests on literature values rather than on measurements from these specimens, its assumed orientation dependence was examined directly. The design motivation in Section 5.1 concerns components held at elevated temperature, so the measurement was made at 650 °C as well as at room temperature. Neutron-diffraction work on IN718 reports that stiffness falls with temperature, from about 220 GPa at room temperature to 140 GPa at 700 °C, while the degree of elastic anisotropy increases, and that the elastic constants are essentially unchanged by ageing [54]. The orientation dependence should therefore persist at service temperature rather than diminish, and any change in it would not be attributable to precipitation state.

The same specimens were tested in tension at room temperature and at 650 °C, so that texture and mechanical response were obtained from the same bar (Table 9). Apparent build-direction stiffness fell from 57 to 29 GPa at room temperature as the measured enrichment increased from 1.15 to 8.38 MUD, and from 46 to 28 GPa over the same specimens at 650 °C; the rank correlations with enrichment were −0.85 and −0.78. The ordering persisted within fixed focus-offset subsets, and partial rank correlations controlling for focus offset or volumetric energy density gave −0.79 and −0.86. The direction of the association is the one assumed in Equation 28, and it holds at both temperatures. The 0.2% proof stress did not follow the enrichment at room temperature, consistent with strength being set principally by the precipitation state rather than by orientation.

**Table 9.** Measured ⟨001⟩ ∥ BD enrichment and tensile response of the nine specimens, ordered by enrichment. Apparent stiffness was obtained from crosshead displacement and includes machine and fixture compliance; it is interpreted as a relative measure only. The last row gives Spearman rank correlations with the measured enrichment. Stiffness follows the enrichment ordering at both temperatures, whereas the proof stress at room temperature does not.

| Condition (P-S-H-F) | $e_{001}$ | $E_{app}$ RM | $E_{app}$ 650 | YS RM | YS 650 |
|---|---|---|---|---|---|
| 400-400-150-0 | 3.49 | 52.0 | 46.1 | 614 | 506 |
| 400-400-250-0 | 2.99 | 53.1 | 33.8 | 598 | 540 |
| 400-600-150-0 | 6.47 | 35.1 | 32.9 | 616 | 507 |
| 400-400-150-(-80) | 4.92 | 40.8 | 29.0 | 584 | 496 |
| 400-400-250-(-80) | 1.15 | 57.1 | 37.4 | 655 | 549 |
| 400-600-150-(-80) | 7.04 | 36.4 | 32.7 | 626 | 503 |
| 400-400-150-(+20) | 7.43 | 37.2 | 30.1 | 612 | 476 |
| 400-400-250-(+20) | 7.07 | 36.3 | 31.2 | 574 | 496 |
| 400-600-150-(+20) | 8.38 | 29.0 | 27.7 | 634 | 507 |
| Spearman ρ with $e_{001}$ | — | -0.85 | -0.78 | -0.02 | -0.58 |

Strain was estimated from crosshead displacement normalized by the nominal gauge length, and the specimens were mounted through threaded end sections, so the apparent values include machine and fixture compliance. Following the recommendation of the collaborators who performed the tests, they are interpreted as a semi-quantitative measure of relative trend among processing conditions rather than as elastic moduli. The observed variation may also reflect differences in porosity or grip conditions that were not measured.

Two constraints apply to the separately built set. Three of its conditions used a negative focus offset, which lies outside the deployment domain of the present study; the validity criterion depends on the magnitude of the focus offset and not its sign, so those conditions were flagged by the applicability weighting rather than by the validity threshold. One condition reached an enrichment of 8.38 MUD, above the value at which the mapping saturates, so its mapped modulus lies at the lower bound in Figure 9.

## 6. Conclusions

This study developed a framework for predicting ⟨001⟩ ∥ build direction (BD) texture intensity with uncertainty quantification in laser powder bed fusion (LPBF). The framework contains two stages: A frozen Stage-1 surrogate mapped process parameters to melt-pool geometry, and Stage 2 combined a conduction-inspired P-V-$\hat{D}$ greybox anchor with residual learning, continuous applicability attenuation, physics-validity abstention, and retained-set empirical uncertainty quantification.

Under leave-one-group-out (LOGO)grouped generalization, the gated hybrid reached $R^2$ of 0.592, compared with 0.538 for the black-box random forest (RF) model. When the complete +80 mm regime was withheld, the greybox and gated hybrid retained $R^2$=0.778 and 0.750, respectively, whereas the black-box RF reached −0.001. The empirical anchor was therefore the principal source of transferable predictive skill in this evaluation. At focus-zero, all texture models produced negative $R^2$, and all 54 queries exceeded the adopted physics-validity threshold, so predictions and intervals were withheld. For issued predictions, the empirical 90% intervals achieved 92.9% coverage at 80.6% retention under LOGO and 100% coverage at full retention for the 42 +80 mm observations. These retained-set values do not constitute a general coverage guarantee under regime shift. The depth ablation associated the +80 mm transfer with the Stage-1 geometry mediator, while Shapley additive explanation (SHAP) and Sobol analyses suggested nonlinear and interaction-dependent hatch behavior outside the parsimonious greybox. Pronounced texture enrichment was observed within conduction-mode (CM) routed conditions, including the +80 mm subset, although the available data do not support a comparative CM versus keyhole-mode (KM) defect claim. A first check on a separate build is reported in Section 5. Broader validation should extend to additional machines and to strain measured with an extensometer, with matched melt-pool, defect, and mechanical-property data. Reliable use under distribution shift therefore required a transferable anchor, controlled residual correction, explicit validity assessment, and transparent retained-set uncertainty rather than model complexity alone.

## CRediT authorship contribution statement

**Yisheng Lu:** Conceptualization, Methodology, Software, Formal analysis, Data curation, Writing – original draft, Writing – review & editing, Visualization. **John Riris:** Investigation, Resources, Data curation, Writing – review & editing. **Jie Song:** Conceptualization, Methodology, Resources, Supervision, Writing – review & editing, Funding acquisition. **Yao Fu:** Conceptualization, Methodology, Resources, Supervision, Writing – review & editing, Funding acquisition. **Jie Chen:** Conceptualization, Methodology, Supervision, Writing – review & editing, Funding acquisition, Project administration.

## Declaration of Competing Interest

The authors declare that they have no known competing financial interests or personal relationships that could have appeared to influence the work reported in this paper.

## Acknowledgments

Y.L. and J.C. acknowledge the support from the startup fund, the 4-VA grant, and the graduate assistantship provided by Virginia Tech, and the Curriculum Development Support provided by MathWorks. Y.F. acknowledges support from the National Science Foundation (Award No. 2104941 and Award No. 2245107) for providing financial support. This work utilized the Nanoscale Characterization and Fabrication Laboratory, a part of the National Nanotechnology Coordinated Infrastructure (NNCI), funded by NSF (ECCS 1542100 and ECCS 2025151).

## Data availability

The processed data and analysis code supporting this study are available from the corresponding author upon reasonable request. These comprise the finalized Stage-2 target table, frozen Stage-1 melt-pool predictions, saved per-observation Stage-2 predictions, uncertainty and mechanism-analysis outputs, and the associated analysis scripts. The target table includes the orientation-point fractions and normalized $e_{001}$ values, and recomputation agrees within 0.003 MUD, reflecting stored decimal precision. The original EBSD orientation maps and the associated crystallographic-processing information are available on the same basis, subject to approval from the data owners.

# Supplementary Information for Physics-based Prediction, uncertainty quantification and decision-making for IN718 crystallographic texture intensity across LPBF defocus regimes

Yisheng Lu, John Riris, Jie Song, Yao Fu, Jie Chen

This document provides the sampling and surrogate settings used for the residual-model interpretation analyses, the complete applicability-gate sensitivity sweep, an independent verification of the texture-target reference fraction, the correspondence between the manuscript group labels and the acquisition identifiers used in the distributed project archive, a scope-aware comparison with representative texture-prediction frameworks, mean absolute percentage errors for the four point predictors, and protocol-matched baselines constructed from literature-derived model forms. All are referenced from the main text. All calculations used deterministic settings and fixed random seeds.

## S1. Sampling and surrogate settings for the residual-model interpretation

Residual-model interpretation was carried out at the controllable-process level using Shapley **additive explanations (SHAP) and Sobol sensitivity analysis. The four controllable process** variables （laser power P, scan speed V, hatch spacing H, and focus offset F) were treated as independent inputs. The two Stage-1 melt-pool mediators required by the six-feature residual model were supplied by deterministic surrogates so that the analysis could be aggregated at the process level rather than at the feature level.

For each sampled process vector, the depth and width mediator surrogates approximated the frozen Stage-1 outputs from P, V, and F; the six-feature residual model was then evaluated on the resulting feature vector. The mediator surrogates were fitted to the frozen Stage-1 outputs available for the 278 Stage-2 observations and were not evaluated on an independent sample. This forward mapping was the target of both the Sobol and the SHAP analyses.

**Table S1.** Sampling and surrogate settings for the SHAP and Sobol analyses. RF denotes random forest.

| Setting | Value | Note |
|---|---|---|
| Sobol sampling design | Saltelli | SALib implementation |
| Sobol base sample size N | 2,048 | 20,480 model evaluations for four inputs |
| Attribution-surrogate training sample | 4,000 | uniform samples over the common bounds |
| TreeSHAP explanation subset | 1,500 | first 1,500 of the same 4,000 samples |
| Residual model | RF, 400 trees | full-data six-feature residual model used for interpretation |
| Depth mediator surrogate | RF, 300 trees | inputs P, V, F; target Stage-1 predicted depth |
| Width mediator surrogate | RF, 300 trees | inputs P, V, F; target Stage-1 predicted width |
| Attribution surrogate | RF, 300 trees | fitted to the residual forward mapping |
| Random seed | 42 | applied to the uniform SHAP sampling and RF fitting; the Saltelli design is deterministic |

*Both analyses used a common product-uniform reference distribution spanning the observed marginal bounds of the four process variables.*

**Table S2.** Uniform sampling bounds used for the SHAP and Sobol reference distribution.

| Process variable | Sampling bounds | Basis |
|---|---|---|
| Laser power, P | 150-400 W | observed marginal range |
| Scan speed, V | 100-1600 mm $s^{-1}$ | observed marginal range |
| Hatch spacing, H | 25-500 µm | observed marginal range |
| Focus offset, F | 0-80 mm | observed marginal range |

Because the rectangular reference distribution assumes independence among the four process variables, the resulting indices describe the fitted residual forward mapping under a synthetic input distribution rather than a variance decomposition of the experimental design. The attribution surrogate was used only to obtain TreeSHAP values over the sampled input space; its goodness of fit was not evaluated on an independent sample, and no generalization claim is made. Because the SHAP and Sobol analyses used the same reference bounds, their agreement represents consistency under a shared model-evaluation distribution rather than independent validation.

## S2. Applicability-gate sensitivity sweep

The deployed applicability gate used a neighborhood size of k = 5 and a reference percentile of 95. Both settings were inherited from canonical implementation and were held fixed throughout the final analysis. The sweep reported here characterizes model behavior around that frozen configuration and was not used to select it.

Each sweep entry was refitted within the applicable training partition, so no held-out texture target contributed to the corresponding gate calibration. One quantity was varied at a time while the other was held at its deployed value.

**Table S3.** Applicability-gate sensitivity sweep. Grouped generalization is reported as leave-one-group-out (LOGO) $R^2$; regime transfer is reported as $R^2$ for the controlled withheld +80 mm evaluation. The deployed configuration is k = 5 with a reference percentile of 95.

| Swept quantity | Value | LOGO $R^2$ | +80 mm $R^2$ | Configuration |
|---|---|---|---|---|
| Neighborhood size k | 3 | 0.592 | 0.759 | |
| | 5 | 0.592 | 0.750 | deployed |
| | 10 | 0.583 | 0.715 | |
| | 15 | 0.577 | 0.656 | |
| Reference percentile | 90 | 0.587 | 0.762 | |
| | 95 | 0.592 | 0.750 | deployed |
| | 99 | 0.574 | 0.606 | |

Grouped generalization varied by less than 0.02 in $R^2$ across the swept neighborhood sizes and reference percentiles. Regime transfer to the withheld +80 mm condition was more sensitive: increasing the neighborhood size from 5 to 15 or the reference percentile from 95 to 99 was associated with lower +80 mm performance, reducing $R^2$ from 0.750 to 0.656 and 0.606, respectively. Smaller neighborhoods and a more restrictive reference percentile produced marginally higher transfer scores but were not adopted, because the deployed configuration was fixed before this sweep was performed.

The smallest evaluated neighborhood size was $k = 3$. Smaller neighborhoods were not examined because the distance estimate becomes increasingly sensitive to individual nearest neighbors as $k$ decreases, and at $k = 1$ the normalized distance is determined entirely by a single training observation, losing its intended interpretation as a local data-density measure.

## S3. Texture-target reference fraction

The texture target was normalized by an adopted random-texture reference fraction of 10.21%, inherited from the original crystallographic-processing workflow. Its consistency with a uniform orientation distribution was evaluated independently.

For a uniformly oriented cubic crystal, the build direction lies within 15° of a symmetry-equivalent ⟨001⟩ direction when it falls inside one of six spherical caps centered on the ± [100], ± [010] and ± [001] poles. Because these poles are separated by 90°, the caps do not overlap at the adopted tolerance, giving the combined solid-angle fraction

$$6\cdot\frac{2\pi(1-\cos 15°)}{4\pi}=3(1-\cos 15°)=10.222\%.$$

An independent Monte Carlo calculation using $5 \times 10^7$ Haar-uniform random rotations, sampled by the Shoemake method with a fixed seed, gave 10.2218% ± 0.0086 percentage points (two binomial standard errors), in agreement with the analytical expectation to 0.0005 percentage points. This calculation does not assume non-overlapping caps and therefore verifies the analytical derivation independently. The script and its output are included in the distributed archive.

The archived reference value differs from the analytical value by 0.012 percentage points, or 0.12% relative. Replacing 10.21% with 10.222% would uniformly rescale $e_{001}$ by a factor of 0.9988, shifting the observed maximum from 7.965 to 7.955 MUD. The archived value was retained to preserve consistency with the finalized target table and all downstream results.

## S4. Experimental group labeling

The nine Stage-2 experimental groups are designated G1 to G9 in the main text. The distributed project archive uses the original acquisition identifiers listed in Table S4. The two labeling systems refer to the same nine groups and the same 278 observations.

**Table S4**. Correspondence between the manuscript group labels and the acquisition identifiers used in the distributed project archive.

| Manuscript label | Acquisition identifier | Observations |
|---|---|---|
| G1 | OR-4 | 38 |
| G2 | OR-5 | 36 |
| G3 | OR-6 | 41 |
| G4 | OR-7 | 30 |
| G5 | OR-8 | 42 |
| G6 | OR-9 | 18 |
| G7 | OR-10 | 21 |
| G8 | OR-11 | 11 |
| G9 | OR-12 | 41 |
| Total | | 278 |

The acquisition identifiers begin at OR-4 because the Stage-2 texture campaign sampled a conduction-focused subset of the earlier melt-pool campaign. Groups OR-1 to OR-3 consisted predominantly of negative focus-offset conditions and keyhole-mode tracks and were not advanced to the main EBSD characterization, since the texture study targets the conduction-dominated regime in which build-direction columnar growth is stable.

**S5. Comparison with representative texture-prediction frameworks**

Reported accuracies from previous LPBF texture-prediction studies follow different target definitions, datasets, and validation designs, and are therefore not directly comparable with the present results. Table S5 summarizes each study according to what it reported rather than placing the studies on a common accuracy scale.

**Table S5.** Scope-aware comparison with representative LPBF crystallographic-texture and microstructure prediction frameworks. NR indicates a quantity not reported by the source study. Metrics retain the target definitions and validation designs of their respective studies and should not be interpreted as like-for-like model rankings

| Study | Material and target | Framework | Evaluation | Reported $R^2$ | Reported MAPE | Other reported performance | Computational pathway | Applicability treatment |
|---|---|---|---|---|---|---|---|---|
| Sofras et al. [21] | 304L; diffraction ratios $r_{220}$ and $r_{200}$ | Decision-tree regression | Ten-fold cross-validation for tree pruning; six newly fabricated control conditions | NR | NR | Training and cross-validation errors; measured versus predicted control values | Trained-tree inference; runtime NR | Dense-processing-window screening; no formal prediction abstention |
| Whitney et al. [22] | Ti-6Al-4V; β-grain morphology and texture, α/α′ descriptors | FDMC–PF–ML hybrid | Five-fold cross-validation and held-out testing for the ML surrogate; FDMC β-texture compared with EBSD at three energy conditions | NR | NR | $R^2 \geq 0.93$ for surrogate phase-fraction and lath-width outputs; not a texture metric | FDMC simulation retained; ML replaces PF response; runtime NR | Temperature-rule routing and interpolation; no query-level output withholding |
| This work | IN718; $e_{001}$ | Physics-anchored gated hybrid | LOGO and complete withheld-defocus evaluations | 0.592 (LOGO); 0.750 (+80 mm) | 27.7% (LOGO); 14.4% (+80 mm) | MAE, RMSE, Spearman ρ, coverage, retention | Trained surrogate inference; no per-query numerical solver; runtime not benchmarked | Residual attenuation and anchor-validity abstention |

**Table S6.** Mean absolute percentage error (MAPE, %) for the four point predictors under the three evaluation settings, calculated as $100\ n^{-1}\ \Sigma\ |\hat{y}_i - y_i|\ /\ y_i$ from the same per-observation predictions used for Table 3. MAPE is interpreted together with the scale-preserving MAE and RMSE values in Table 3 because it assigns greater weight to observations with lower $e_{001}$; only one observation had $e_{001} < 1.0$ MUD. Focus-zero values are diagnostic because no operational prediction was issued for this subset.

| Evaluation setting | Greybox anchor | Black-box RF | Ungated hybrid | Gated hybrid |
|---|---|---|---|---|
| LOGO generalization | 35.4 | 28.9 | 28.5 | 27.7 |
| +80 mm withheld | 13.6 | 27.9 | 27.8 | 14.4 |
| Focus-zero withheld | 59.5 | 85.1 | 66.7 | 61.5 |

## S6. Protocol-matched baselines from literature-derived model forms

Whereas Table S5 compares the reported scope of representative studies, this section evaluates literature-derived model forms numerically on the present dataset. These comparisons were constructed post hoc as diagnostic benchmarks and were not used for model selection or framework configuration. Each entry is an implementation of a published procedure or descriptor applied to the present IN718 dataset, not a reproduction of published results: the material, the target definition and the validation design all differ from the source studies. Every comparator followed its intended or frozen configuration, with all data-dependent tuning confined to the outer-training partition.

**Table S7.** Literature-derived model forms evaluated on the present dataset under the same LOGO and withheld-defocus protocols used throughout. The reduced geometric surrogate retains only a single-pool boundary-orientation calculation and omits the multi-track and multi-layer growth selection implemented in the full mechanistic model, so it is not a reproduction of that model. Focus-zero values are diagnostic; see below. MAPE for these baselines is not reported because several give negative $R^2$ and their relative errors are not informative

| Model form | Source basis | Implementation status | LOGO $R^2$ | +80 mm $R^2$ | Focus-zero $R^2$ |
|---|---|---|---|---|---|
| Regression tree, P V H | Sofras et al. [21] | source-method reimplementation | 0.191 | 0.175 | −1.278 |
| Regression tree, P V H F | Sofras et al. [21] | dataset-motivated adaptation | 0.302 | 0.297 | −0.961 |
| Regression tree, P V H + VED | Sofras [21]; Caiazzo et al. [47] | dataset-motivated adaptation | 0.241 | 0.112 | −1.186 |
| Volumetric energy-density power law | Caiazzo et al. [47] | literature-derived descriptor baseline | −0.065 | −0.383 | −3.930 |
| Linear energy-density power law | Caiazzo et al. [47] | literature-derived descriptor baseline | −0.038 | −0.517 | −2.494 |
| Normalized enthalpy, IN718 properties | King et al. [12] | dataset-motivated adaptation | −0.203 | 0.028 | 0.002 |
| Reduced melt-pool geometric surrogate | Liu et al. [48] | reduced literature-inspired surrogate | −0.170 | 0.013 | −0.194 |
| Greybox anchor | — | this work | 0.287 | 0.778 | −0.298 |
| Black-box random forest | — | this work | 0.538 | −0.001 | −1.092 |
| Ungated hybrid | — | this work | 0.564 | 0.199 | −0.770 |
| Gated hybrid | — | this work | 0.592 | 0.750 | −0.378 |

The focus-zero column is reported for diagnostic purposes only. The literature-derived baselines and the internal point-prediction baselines issue a prediction for every query and implement no abstention rule, whereas the proposed framework withholds all focus-zero predictions. A negative diagnostic coefficient for the gated hybrid therefore does not indicate an operational prediction failure.

| **Model class** | **Focus-zero treatment** |
| --- | --- |
| Literature-form baselines | Diagnostic prediction; no abstention rule |
| Internal point-prediction baselines | Diagnostic prediction |
| Proposed framework | Prediction withheld |

**Table S8.** Sensitivity of the +80 mm coefficient of determination to the cost-complexity pruning implementation. The canonical implementation computes the exact pruning path on each training partition and selects the subtree with minimum mean ten-fold validation error, following the published description most closely; the alternatives use pre-specified alpha grids. Across all three implementations the defocus-aware tree remained below the physics anchor (0.778).

| Predictor set | Canonical (exact pruning path) | Fixed log grid | Fixed linear grid |
| --- | --- | --- | --- |
| P V H | 0.175 | 0.130 | 0.138 |
| P V H F | 0.297 | 0.297 | 0.325 |

## S7. Reproducibility

All analyses used deterministic settings. The applicability-gate sweep was regenerated through leakage-free refitting within each evaluation partition, whereas the interpretation analyses were regenerated from the finalized Stage-2 table and the frozen Stage-1 mediator outputs. Canonical per-observation predictions and saved mechanism-analysis outputs were used for verification. The literature-derived baselines in Section S6 were regenerated by refitting within each evaluation partition, with all pruning selection confined to the outer-training partition; per-observation predictions for every predictor set and protocol are included in the archive. The distributed project archive contains the associated scripts and intermediate files.